\documentclass{article}

\usepackage{microtype}
\usepackage{graphicx}
\usepackage{subcaption}
\usepackage{booktabs} 

\usepackage{hyperref}

\PassOptionsToPackage{table}{xcolor}  
\usepackage{xcolor}                   
\usepackage{colortbl}  
\usepackage[preprint]{icml2026}

\usepackage{amsmath}
\usepackage{amssymb}
\usepackage{multirow}
\usepackage{mathtools}
\usepackage{amsthm}

\usepackage[capitalize,noabbrev]{cleveref}

\theoremstyle{plain}

\theoremstyle{definition}

\theoremstyle{remark}

\usepackage[textsize=tiny]{todonotes}

\newcommand{\rebuttal}[1]{\textcolor{blue}{#1}}

\renewcommand{\rebuttal}[1]{#1}

\newcommand{\methodname}{{R-Stitch}}

\begin{document}

\twocolumn[
  \icmltitle{R-Stitch: Dynamic Trajectory Stitching for Efficient Reasoning}

  \icmlsetsymbol{corresponding}{$\dagger$}

  \begin{icmlauthorlist}
    \icmlauthor{Zhuokun Chen}{monash}
    \icmlauthor{Zeren Chen}{buaa}
    \icmlauthor{Jiahao He}{zju}
    \icmlauthor{Lu Sheng}{buaa}
    \icmlauthor{Mingkui Tan}{scut}
    \icmlauthor{Jianfei Cai}{monash}
    \icmlauthor{Bohan Zhuang}{zju,corresponding}
  \end{icmlauthorlist}

  \icmlaffiliation{monash}{Monash University, Australia}
  \icmlaffiliation{buaa}{School of Software, Beihang University, China}
  \icmlaffiliation{scut}{South China University of Technology, China}
  \icmlaffiliation{zju}{ZIP Lab, Zhejiang University, China}

  \icmlcorrespondingauthor{Zhuokun Chen}{caesard216@gmail.com}
  \icmlcorrespondingauthor{Bohan Zhuang}{bohan.zhuang@gmail.com}

  \icmlkeywords{Machine Learning, ICML}

  \vskip 0.3in
]



\printAffiliationsAndNotice{}  

\begin{abstract}
Chain-of-thought (CoT) improves the reasoning ability of large language models (LLMs) but incurs substantial inference cost due to long autoregressive trajectories.
Existing acceleration methods either shorten reasoning traces or rely on speculative decoding with a smaller model.
However, speculative decoding offers limited speedup when model agreement is low and rigidly enforces token-level consistency, failing to exploit the fact that smaller models often produce much shorter reasoning traces when correct.
We introduce \emph{R-Stitch}, a training-free hybrid decoding framework that delegates token-level computation between a small language model (SLM) and an LLM using entropy as an uncertainty proxy.
High-uncertainty tokens are routed to the LLM, while low-uncertainty tokens are handled by the SLM, avoiding full rollbacks and preserving answer quality.
We further propose \emph{R-Stitch$^+$}, which learns an adaptive routing policy to dynamically adjust token budgets beyond fixed thresholds.
By jointly reducing per-token decoding cost and the number of generated tokens, our approach achieves substantial acceleration with negligible accuracy loss.
It attains peak speedups of $3.00\times$ on DeepSeek-R1-Distill-Qwen-7B, $3.85\times$ on 14B, and $4.10\times$ on QWQ-32B, while maintaining accuracy comparable to full LLM decoding.
Moreover, R-Stitch enables flexible efficiency--accuracy trade-offs under diverse computational budgets without retraining.
Project is available at \href{https://caesarhhh.github.io/R-Stitch}{https://caesarhhh.github.io/R-Stitch}.
\end{abstract}

%


\section{Introduction}

Large language models (LLMs) have achieved impressive performance on a wide range of reasoning tasks, particularly when combined with chain-of-thought (CoT) prompting~\citep{wei2022chain,zhang2024chain,zheng2024critic}. By generating intermediate reasoning steps token-by-token, CoT enables LLMs to solve complex problems in arithmetic, logic, and code generation~\citep{wei2022chain,kojima2022large,zhang2210automatic}. However, this autoregressive decoding process is inherently slow, as each token requires a full forward pass through the model~\citep{liu2024kangaroo,sadhukhan2024magicdec,chen2023accelerating}. The latency becomes especially problematic in CoT, where outputs often span thousands of tokens, significantly limiting the applicability of LLMs in time-sensitive scenarios. 

To address this bottleneck, recent efforts have primarily focused on three directions: reducing the number of generated tokens, applying speculative decoding to accelerate the generation process, and optimizing KV cache access during long-context decoding. The first direction includes approaches that aim to shorten CoT sequences, such as early exiting or incorporating length-aware reward functions during Reinforcement Learning (RL)~\citep{fatemi2025concise,yang2025dynamic,ma2025reasoning,yi2025shorterbetter,jiang2025think}. The second leverages a small language model (SLM) to draft multiple tokens ahead, which are then verified in parallel by a larger LLM~\citep{liu2024kangaroo,sadhukhan2024magicdec,chen2023accelerating}. If verification succeeds, the drafts are accepted; otherwise, generation rolls back to the last matching token. The third direction addresses the I/O bottleneck introduced by repeated access to large key-value (KV) caches during decoding, which becomes increasingly expensive in long-context reasoning. Techniques such as sparse or selective KV caching~\citep{gao2025seerattentionr,tang2024questqueryawaresparsityefficient} have been proposed to reduce memory traffic and improve decoding speed on modern hardware.

\begin{figure*}[t]
    \centering
    \begin{subfigure}[b]{0.325\textwidth} 
        \includegraphics[width=\textwidth]{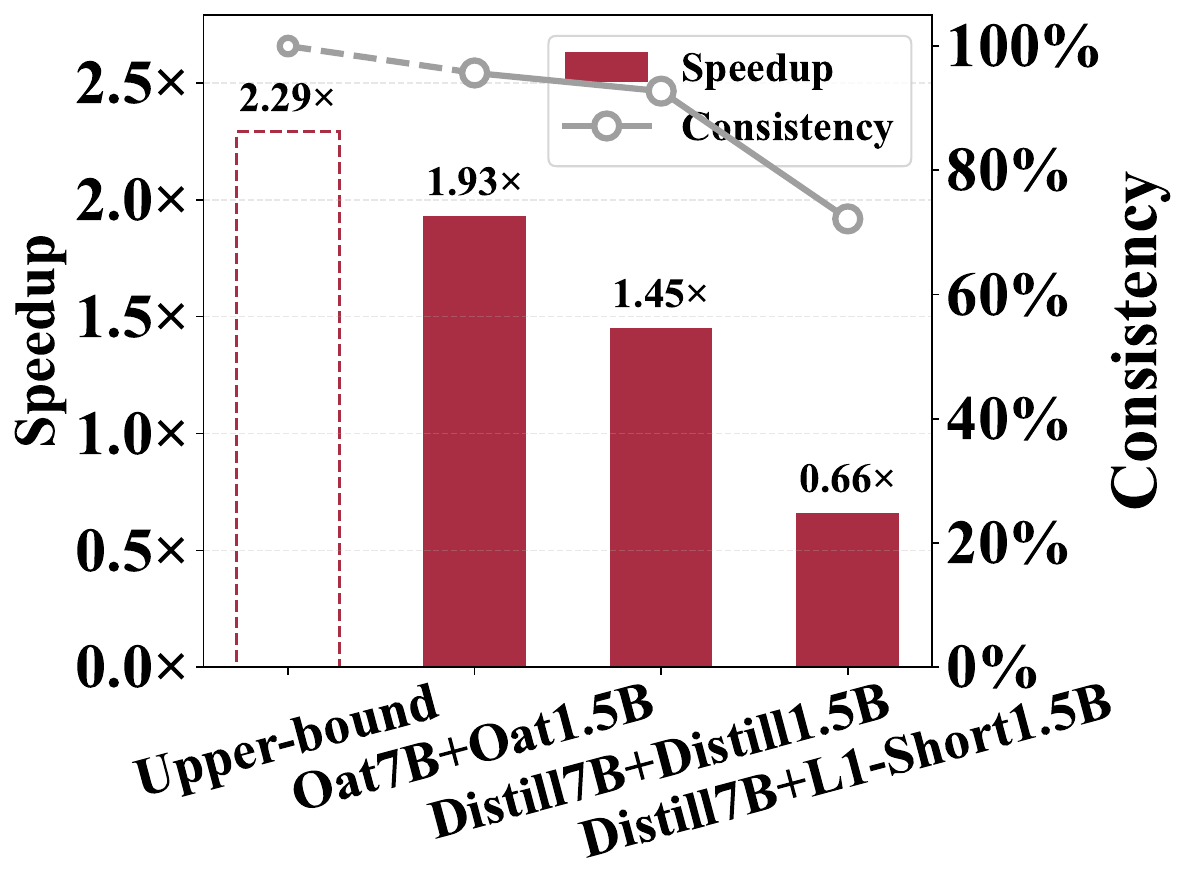}
        \caption{Token-level consistency versus speedup across different LLMs.}
        \label{fig:consistency_dllms}
    \end{subfigure}
    \begin{subfigure}[b]{0.325\textwidth} 
        \includegraphics[width=\textwidth]{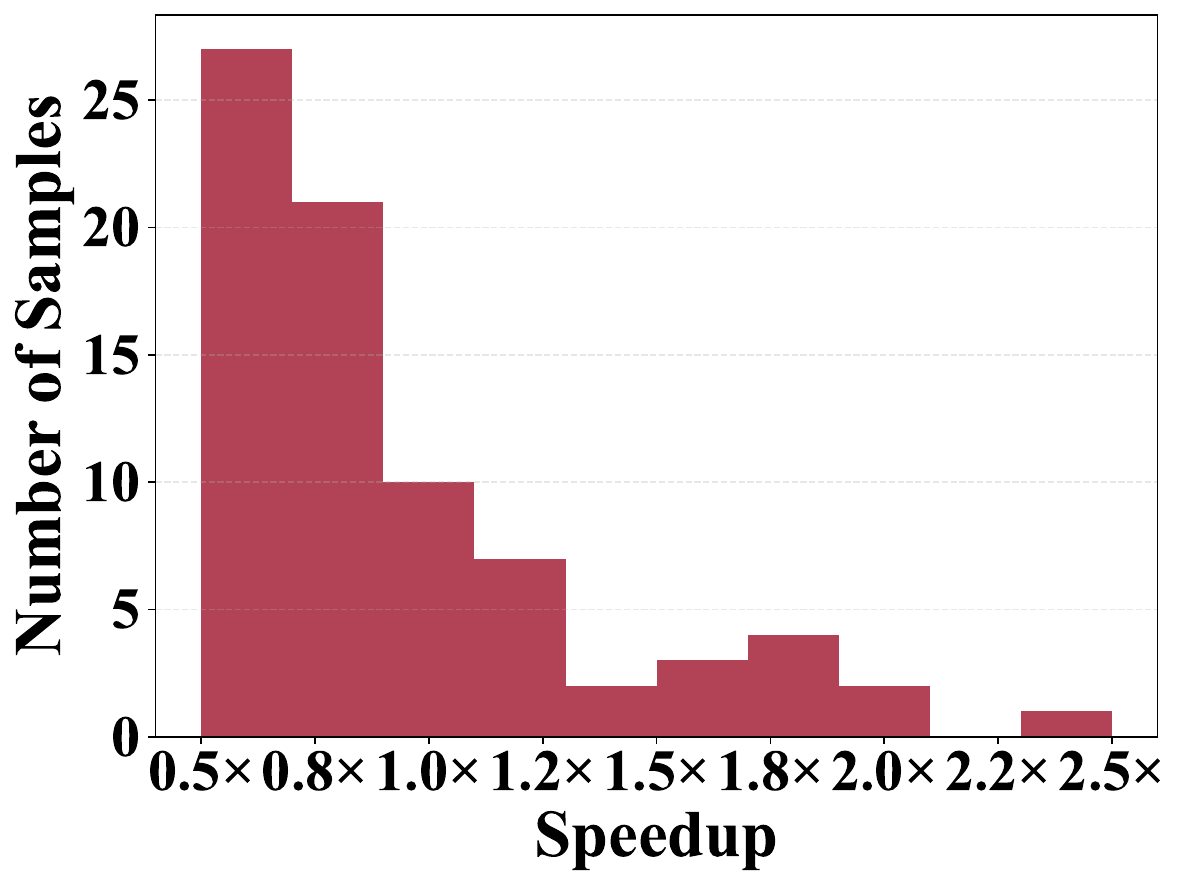}
        \caption{Speedup across individual samples in AMC.}
        \label{fig:consistency_dsamples}
    \end{subfigure}
    \begin{subfigure}[b]{0.325\textwidth} 
        \includegraphics[width=\textwidth]{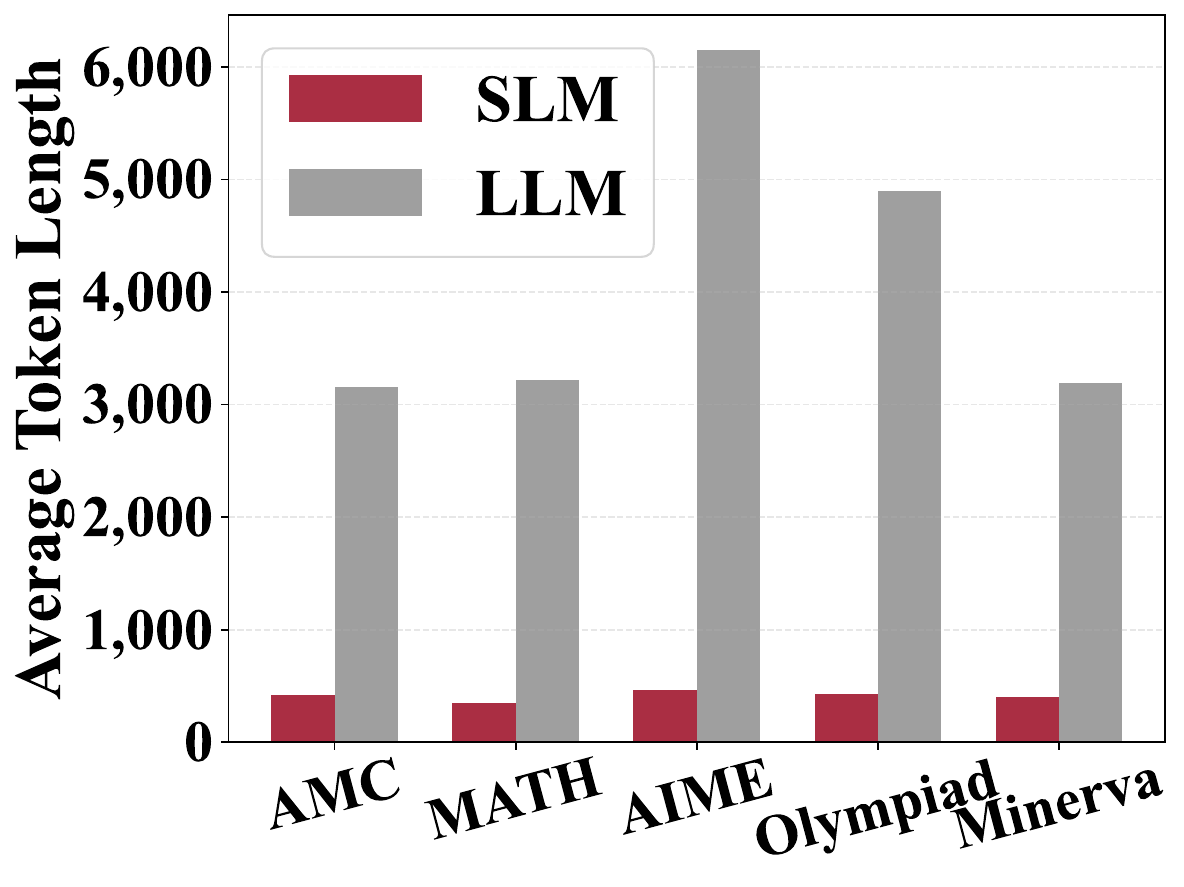}
        \caption{Token usage on questions answered correctly by SLM and LLM.}
        \label{fig:token_len}
    \end{subfigure}
    \caption{
\textbf{Token-level consistency and speedup analysis.}
(a) shows the relationship between token-level consistency and speedup in speculative decoding across different LLM-SLM pairs on AMC.  
(b) presents the distribution of speedup ratios across individual samples from AMC.  
(c) illustrates the token counts for questions correctly answered by both the SLM and LLM.
}
    \label{fig:motivation_spec}
\end{figure*}

Among the three directions, speculative decoding has received considerable attention due to its potential for substantial speedups. However, its effectiveness critically depends on the consistency between the small language model (SLM) and the large language model (LLM). We quantify this limitation using token-level consistency, defined as the percentage of tokens for which the SLM produces the same output as the LLM given an identical prefix. Figure~\ref{fig:consistency_dllms} presents token-level consistency and decoding speedup in speculative decoding across four model combinations on the AMC dataset: DeepSeek-R1-Distill-Qwen-1.5B/7B~\citep{guo2025deepseek}, L1-1.5B-Short~\citep{aggarwal2025l1} and Qwen2.5-Math-1.5B/7B-Oat-Zero~\citep{liu2025understanding}. The results show a clear trend: lower consistency correlates with smaller speedup, particularly in reasoning-intensive tasks. Figure~\ref{fig:consistency_dsamples} shows the distribution of speedup ratios across individual samples from the AMC dataset, using DeepSeek-R1-Distill-Qwen-7B as the LLM and L1-Short as the SLM. While some samples achieve noticeable acceleration, a large number have speedups below 1$\times$, indicating that speculative decoding can introduce overhead on certain inputs. Furthermore, Figure~\ref{fig:token_len} compares the number of tokens generated by the same SLM and LLM on questions both models answer correctly. The SLM produces much shorter completions, suggesting that speculative decoding's rigid reliance on exact token agreement may prevent it from utilizing the SLM’s more concise reasoning effectively.

To more flexibly exploit SLM for acceleration, we propose \methodname, an entropy-guided decoding framework inspired by the preceding observations. Our empirical analysis shows that tokens with higher entropy are more error-prone, which motivates a token-level routing strategy: the SLM acts as the primary generator, and the LLM is invoked only when needed. In this framework, confident low-entropy tokens are accepted directly from the SLM, while uncertain high-entropy tokens trigger LLM intervention for correction and continued generation. Compared with agreement-based speculative decoding, this dynamic delegation avoids full-sequence rollbacks, preserves the complementary strengths of both models, and achieves efficient reasoning under tight computational budgets.

To refine the entropy-based heuristic, we introduce R-Stitch$^{+}$. Rather than relying on a fixed threshold, R-Stitch$^{+}$ equips high-uncertainty tokens with a lightweight router that determines whether LLM intervention is necessary. The router is trained via RL with a latency-aware reward, enabling a data-driven policy that adaptively balances accuracy and efficiency. Thus, R-Stitch$^{+}$ extends the heuristic rule of R-Stitch into a more effective routing mechanism under diverse conditions.

We validate our proposed method on five challenging mathematical reasoning benchmarks using DeepSeek-R1-Distill-Qwen models at 7B, 14B, and 32B scales. 
Across all settings, our approach achieves substantial latency reduction while maintaining accuracy close to full LLM decoding, reaching peak speedups of $3.00\times$, $3.85\times$, and $4.10\times$ on the 7B, 14B, and 32B models, respectively. 
Overall, these results confirm that token-level, entropy-guided collaboration provides a principled and effective solution for accelerating CoT reasoning, addressing the limitations of speculative decoding and enabling flexible deployment under different computational budgets.

Our contributions are summarized as follows:
\begin{itemize}

    \item We analyze the limitations of speculative decoding in low-consistency CoT settings and show that its rigid alignment with the LLM can sacrifice potential efficiency gains from the SLM.
    
    \item We analyze the entropy distribution of tokens in CoT reasoning and reveal its strong correlation with prediction errors, motivating entropy as an effective routing signal.
    
    \item We propose R-Stitch, an entropy-guided hybrid decoding paradigm that adaptively switches between the SLM and LLM to accelerate CoT generation without requiring additional training. We further extend it to R-Stitch$^+$, which learns a routing policy with a latency-aware RL reward, enabling more optimal efficiency–accuracy trade-offs.  

    \item Extensive experiments on mathematical reasoning benchmarks demonstrate that our method achieves consistently lower latency at the same level of accuracy compared with speculative decoding.  

\end{itemize}

\section{Related Work}
\label{gen_inst}

\noindent\textbf{LLM reasoning.}
Recent advances in prompting strategies have significantly improved the reasoning capabilities of LLMs, enabling them to solve complex tasks through structured intermediate computation. A variety of inference-time strategies have been proposed to enhance reasoning, primarily including Chain-of-Thought (CoT)~\citep{wei2022chain,zhang2024chain,zheng2024critic}, Tree-of-Thought~\citep{yao2023tree}, and Monte Carlo Tree Search (MCTS)-based decoding~\citep{zhang2023planning}. Among these, CoT has emerged as a widely adopted method that guides the model to verbalize intermediate steps instead of directly predicting the final answer~\citep{wei2022chain,kojima2022large,zhang2210automatic,li2025s}, improving performance on arithmetic and code reasoning benchmarks. However, these approaches often rely on token-by-token autoregressive decoding of long reasoning chains, which can be inefficient and lead to high inference latency, limiting their scalability in real-world applications.

\noindent\textbf{Accelerating reasoning in LLMs.}
Improving the efficiency of chain-of-thought (CoT) reasoning involves addressing three key bottlenecks: reasoning length, per-token decoding cost, and KV cache I/O overhead. The first focuses on reducing the number of generated tokens by shortening reasoning chains. Methods such as Concise Reasoning~\citep{fatemi2025concise} and ShorterBetter~\citep{yi2025shorterbetter} use length-aware rewards to promote brevity without sacrificing correctness. Other approaches, including DEER~\citep{yang2025dynamic} and ThinkPrune~\citep{hou2025thinkprune}, leverage model entropy or attention patterns to eliminate unnecessary steps, while methods like AdaptThink~\citep{zhang2025adaptthink} and ThinkLess~\citep{fang2025thinkless} aim to skip reasoning entirely when it is not needed. These techniques effectively reduce sequence length, thereby lowering total decoding time. The second line of work targets per-token decoding complexity. Speculative decoding~\citep{liu2024kangaroo,sadhukhan2024magicdec,chen2023accelerating} speeds up generation by having a small language model (SLM) propose token drafts, which are verified in parallel by a large LLM. However, low agreement between the SLM and LLM often causes frequent rollbacks, limiting acceleration gains. Finally, recent research ~\citep{gao2025seerattentionr}~\citep{tang2024questqueryawaresparsityefficient}has highlighted the importance of reducing KV cache I/O—an increasingly dominant cost in long-context decoding, which also happens during the inference phase of reasoning models due to the length of CoT. Since decoding is often I/O-bound, reducing the loading of KV cache from global memory to on-chip memory of GPUs (e.g., via sparse or selective KV caching) can significantly improve system throughput.

\noindent\textbf{Stitching.} Stitching broadly refers to composing multiple models or modules into a unified inference process, often by connecting their intermediate representations or generation trajectories. Early approaches like SN-Net~\citep{pan2023stitchable}, SN-Net v2~\citep{pan2024stitched}, and ESTA~\citep{he2024efficient} focused on layer-wise stitching within single-modality models, enabling efficient adaptation across backbone variants. Other works such as T-Stitch~\citep{tstitch} extended stitching to the trajectory level, applying diffusion-based denoising to merge reasoning paths. Beyond single domains, cross-model stitching has also emerged—LLaVA~\citep{liu2023visual}, for example, connects a vision encoder with an LLM to build a multi-modal agent. More recent efforts explore general-purpose stitching frameworks, such as MetaQuery~\citep{pan2025transfer} and UniWorld~\citep{lin2025uniworld}, which introduce connector modules (e.g., MLPs or learned queries) to link heterogeneous models. 
Our work follows this trend by stitching a small and large language model at the token level during autoregressive decoding, enabling efficient collaboration without expensive retraining.


\section{Methodology}

In this section, we build on our empirical finding that high-entropy tokens are more likely to induce errors, and present \methodname, a collaborative decoding framework that uses entropy as an uncertainty proxy to coordinate the SLM and LLM. \methodname~accepts SLM tokens when entropy is low to reduce latency, while delegating high-entropy tokens to the LLM for reliable decoding. This entropy-guided routing exploits the complementary strengths of heterogeneous models and underpins the RL extension \methodname$^+$, which learns an adaptive policy beyond fixed thresholds.

\subsection{Preliminary}

\textbf{Hybrid decoding setting.}  
We study a hybrid inference setup with a SLM \( f_{\text{SLM}} \) and a LLM \( f_{\text{LLM}} \), both autoregressive and sharing the same tokenizer. Given a prompt \( x \), the model generates a token sequence \( y_{1:T} = [y_1, y_2, \dots, y_T] \) in an autoregressive manner. At decoding step \( t \), the active model \( f \in \{f_{\text{SLM}}, f_{\text{LLM}}\} \) outputs a probability distribution \( \mathbf{p}_t = f(y_{1:t-1}) \), from which the next token \( y_t \) is sampled.

\begin{figure*}[t]
  \centering
  \includegraphics[width=\textwidth]{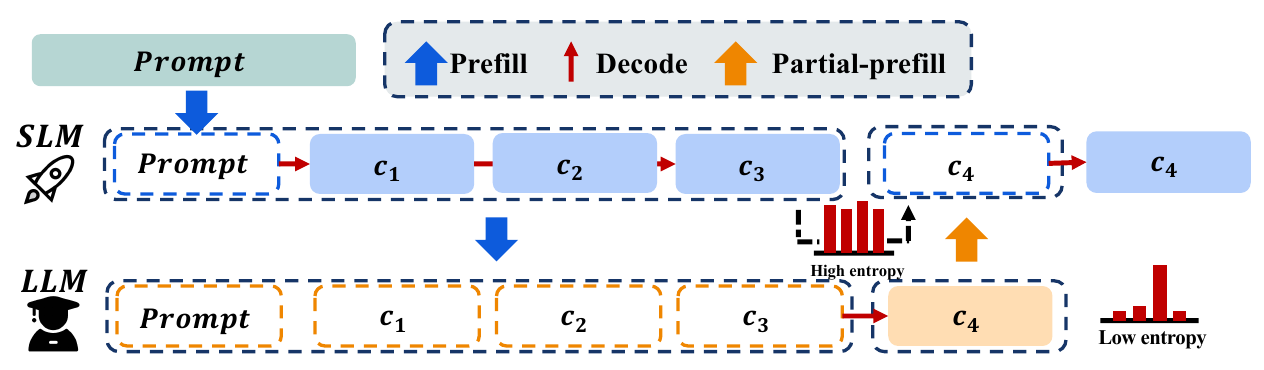}
   \caption{
\textbf{Overview of \methodname.} Given a question with CoT prompting, decoding alternates between an SLM and an LLM under an entropy-based switching policy. Generation starts with the SLM; tokens with low entropy are accepted directly, while high-entropy tokens trigger the LLM to overwrite them and resume decoding. Symmetrically, when the LLM outputs a low-entropy token, it returns to the SLM to reduce computational cost. This bidirectional mechanism adaptively allocates computation, preserving SLM efficiency while leveraging LLM reliability when needed.
}
   \label{fig:radar}
\end{figure*}

\subsection{Bidirectional Entropy-Guided Decoding}

\subsubsection{Empirical Entropy Analysis}

LLMs exhibit stronger reasoning capabilities but incur substantially higher inference costs compared to SLMs.
Prior work shows that when the two models produce consistent outputs, delegating computation to the SLM can effectively reduce latency.
This raises a fundamental question: under what conditions can SLM predictions be relied upon without compromising correctness?
To answer this, we conduct sample- and token-level entropy analyses on AMC using DeepSeek-R1-Distill-Qwen-7B (LLM) and L1-1.5B-Short (SLM), which reveal the following empirical patterns.

\begin{figure*}[t]
    \centering
    \begin{minipage}{0.32\linewidth}
        \centering
        \includegraphics[width=\linewidth]{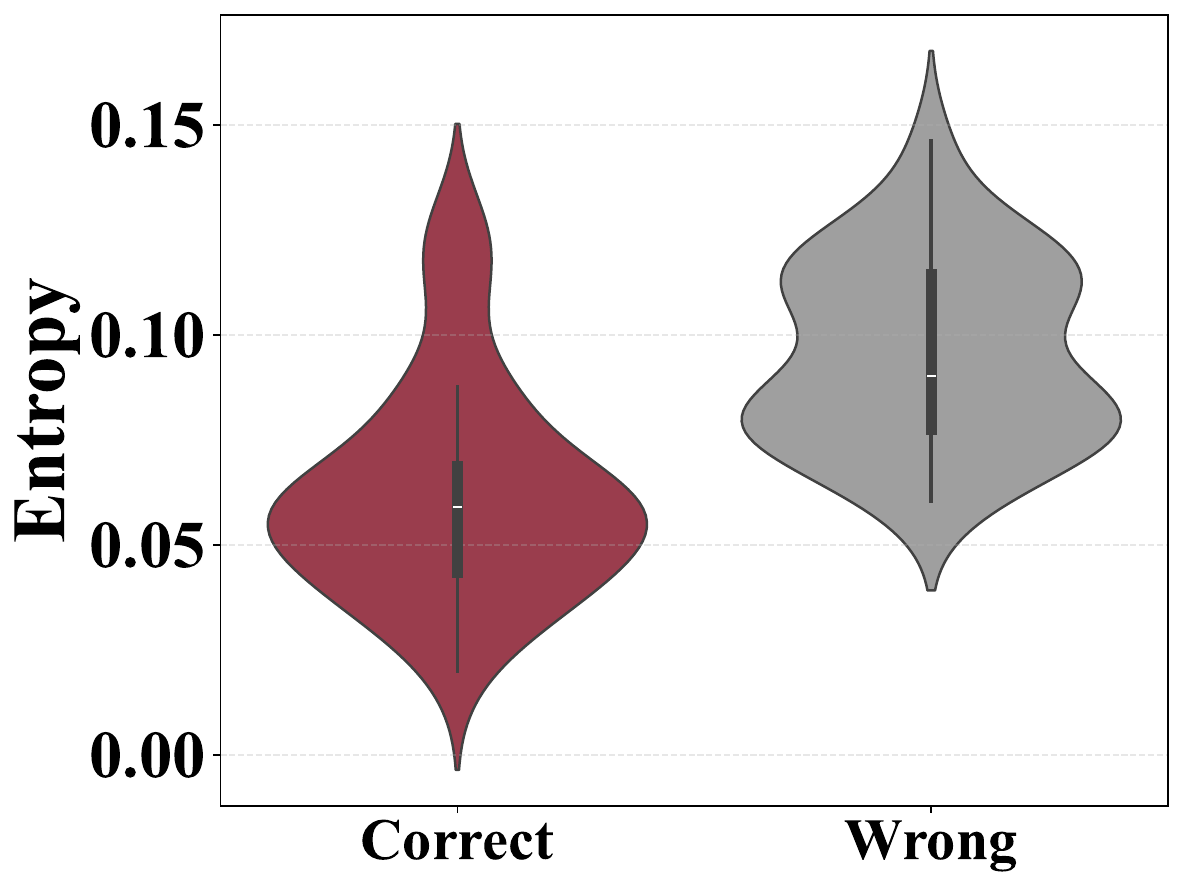}
        \subcaption{Sample-level entropy in correct vs.\ incorrect solutions \label{fig:motivation-a}}
    \end{minipage}
    \hfill
    \begin{minipage}{0.32\linewidth}
        \centering
        \includegraphics[width=\linewidth]{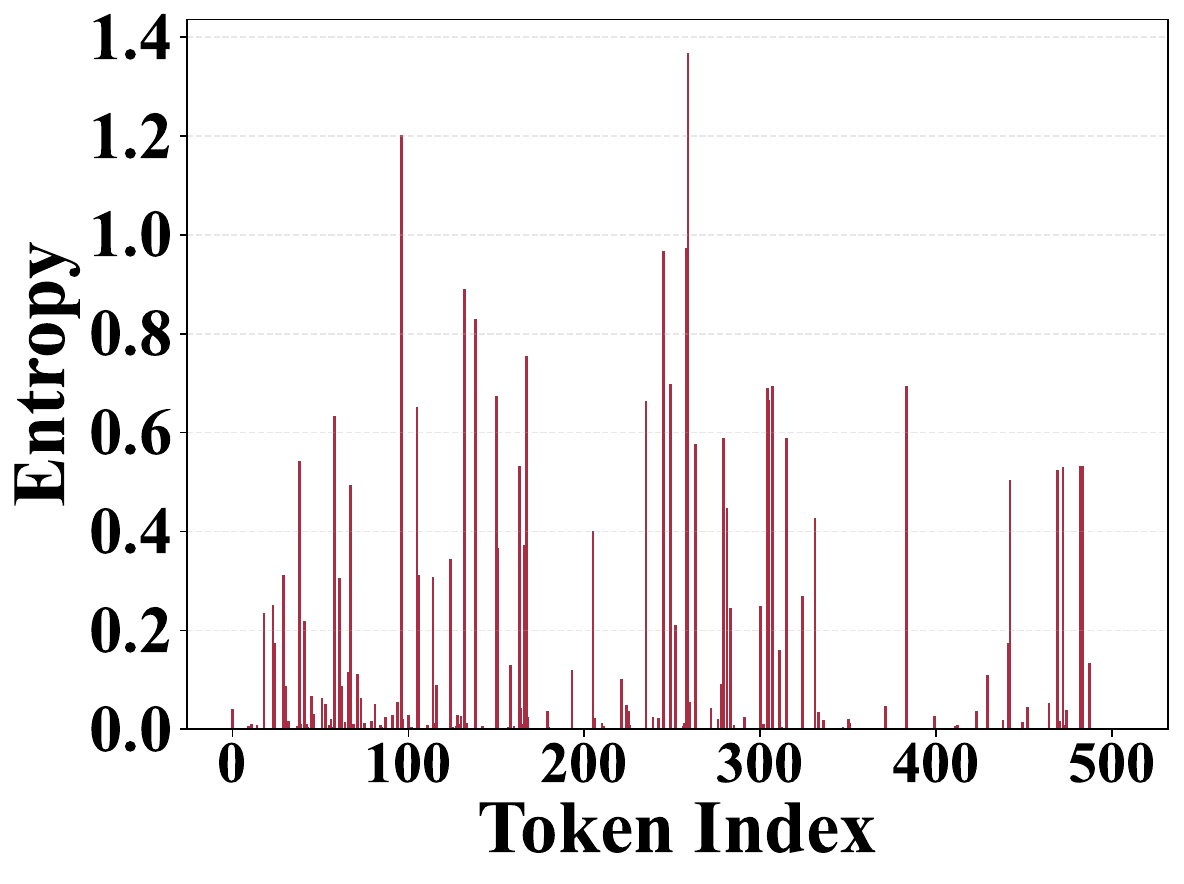}
        \subcaption{Token-level entropy distribution across full reasoning traces\label{fig:motivation-b}}
    \end{minipage}
    \hfill
    \begin{minipage}{0.32\linewidth}
        \centering
        \includegraphics[width=\linewidth]{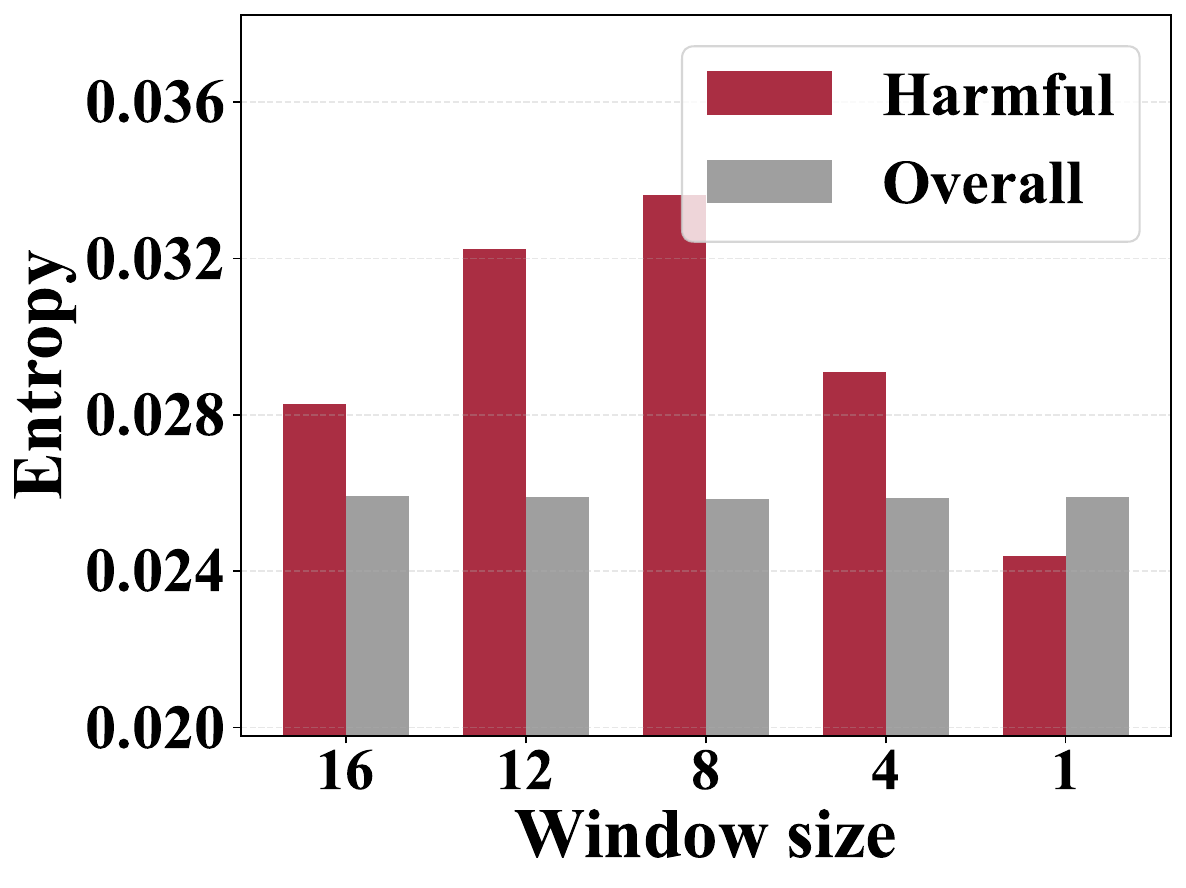}
        \subcaption{Elevated entropy around the first harmful token\label{fig:motivation-c}}
    \end{minipage}
    \caption{
        \textbf{Entropy and error locality.}
        (a) Incorrect solutions exhibit higher entropy than correct ones.
        (b) The entropy distribution is heavily skewed toward zero; most tokens have (near-)zero entropy.
        (c) Neighborhoods around the first harmful token show higher mean entropy than overall entropy distribution, indicating that errors often arise from locally uncertain regions.
    }
    \label{fig:motivation}
\end{figure*}

\textbf{1. Elevated uncertainty is associated with error-prone regions.}
On AMC, the mean entropy over generated tokens is higher in incorrect reasoning traces than in correct ones (Figure~\ref{fig:motivation-a}).
To further localize this effect, we analyze \emph{harmful tokens}, defined as the first SLM-generated tokens whose inclusion flips the LLM’s answer from correct to incorrect, restricting the analysis to cases where the LLM succeeds but the SLM fails.
As shown in Figure~\ref{fig:motivation-c}, the local context preceding such harmful tokens consistently exhibits higher entropy than the overall token distribution, indicating that errors tend to arise in locally uncertain regions.

\textbf{2. Most tokens are generated with extremely low uncertainty.}
We analyze the entropy distribution across reasoning sequences.
As shown in Figure~\ref{fig:motivation-b}, the distribution is heavily skewed toward zero: only 10.65\% of SLM-generated tokens exceed an entropy of 0.1, while the majority have entropy exactly 0.
This shows that high-entropy tokens are relatively rare within a sequence.

\subsubsection{Method}
Building on these findings, we design R-Stitch to explicitly exploit the observed link between token entropy and error likelihood.
R-Stitch is a token-level decoding framework that dynamically alternates between the SLM and LLM based on entropy-derived uncertainty. The key idea is to let the SLM decode as much as possible to reduce latency, while invoking the LLM only when necessary to maintain answer quality. Importantly, the LLM can also transfer control back to the SLM once low-entropy (high-certainty) tokens are reached, enabling a full bidirectional flow. We use a predefined threshold $\tau$ to balance the latency–accuracy trade-off.

\noindent\textbf{Entropy-guided stitching.} At each decoding step $t$, the active model produces a probability distribution $\mathbf{p}_t$ over the vocabulary of size $V$. We quantify uncertainty by the normalized entropy of this distribution:
\begin{equation}
\label{eq:entropy}
\mathcal{H}_t \;=\; \frac{-\sum_{i=1}^{V} p_{t,i}\,\log p_{t,i}}{\log V}, 
\qquad \mathcal{H}_t \in [0,1].
\end{equation}

Larger $\mathcal{H}_t$ indicates higher uncertainty, while smaller values indicate higher certainty.

Decoding begins with the SLM. At each step $t$, if the uncertainty $\mathcal{H}^{\text{SLM}}_t \le \tau$, the SLM accepts its prediction and proceeds. Otherwise ($\mathcal{H}^{\text{SLM}}_t > \tau$), the token is discarded and the step is reprocessed by the LLM, which resumes decoding. Symmetrically, while decoding with the LLM, if $\mathcal{H}^{\text{LLM}}_t \le \tau$, control is handed back to the SLM; otherwise, the LLM continues. 

Formally, the switching policy is:
\begin{equation}
\label{eq:switch_r_stitch}
\text{Switch}(t) =
\begin{cases}
\text{SLM} \rightarrow \text{LLM} & \text{if } \mathcal{H}_t^{\text{SLM}} > \tau, \\
\text{LLM} \rightarrow \text{SLM} & \text{if } \mathcal{H}_t^{\text{LLM}} \le \tau.
\end{cases}
\end{equation}

\noindent\textbf{KV Cache management.} \methodname~maintains separate key-value caches for the SLM and LLM. When a model first participates in reasoning, it performs a full prefill over the existing input context to initialize its KV cache. Thereafter, each model incrementally updates its own cache during decoding. During model switches, we avoid redundant computation by leveraging \emph{partial prefill}. Specifically, when switching back to a model that has previously decoded, we reuse its existing KV cache and only prefill the new tokens generated by the other model since the last switch. This strategy eliminates repeated attention over already-processed tokens, thereby enabling efficient cache reuse and significantly reducing switching overhead.

\subsection{RL-based Routing with Latency-aware Reward}

Beyond entropy-guided stitching, we propose \textbf{R-Stitch$^+$}, an RL extension of R-Stitch. At each decoding step, when the token entropy exceeds the threshold $\tau$, the hidden state of the current model is fed into a lightweight router, which decides whether to continue with the SLM or switch to the LLM. Low-entropy tokens are always delegated to the SLM.

\textbf{Reward design.} The reward is decomposed into an accuracy term and an efficiency term:
\begin{equation}
\label{eq:reward_overall}
R = r_{\text{acc}} + r_{\text{eff}}.
\end{equation}
The accuracy reward $r_{\text{acc}}$ reflects whether the final prediction is correct, while the efficiency reward $r_{\text{eff}}$ penalizes computational cost. Unlike prior RL approaches for chain-of-thought reasoning that approximate efficiency by the number of generated tokens~\citep{aggarwal2025l1}, such a proxy is unsuitable here: (i) we involve both a small and a large model, whose per-token costs differ significantly; and (ii) our framework allows partial-prefill operations, whose cost depends jointly on the KV cache size and the length of the newly prefilling span. Ideally, the most faithful measure of efficiency is the actual trajectory latency, so we define
\begin{equation}
\label{eq:reward_eff}
r_{\text{eff}} = - \lambda \cdot r_{\text{acc}} \cdot \widehat{L},
\end{equation}
where $\widehat{L}$ denotes trajectory latency and $\lambda$ is a trade-off coefficient. Latency is penalized only when the output is correct, ensuring that the router does not pursue speed at the expense of accuracy. 

\begin{table*}[t]
\centering
\caption{
Comparison of decoding strategies on five mathematical reasoning datasets. 
We report accuracy, latency (s/sample), and relative speedup (computed against the corresponding full LLM decoding) under decoding budgets of 8k and 16k tokens. 
\textbf{LLM-7B}, \textbf{LLM-14B}, and \textbf{LLM-32B} denote DeepSeek-R1-Distill-Qwen-7B, DeepSeek-R1-Distill-Qwen-14B, and QWQ-32B, respectively. 
\textbf{SLM} refers to L1-1.5B-Short~\citep{liu2025understanding}, and \textbf{SpecDec} denotes speculative decoding using the corresponding models. 
Speedup is defined per LLM, per budget, per dataset as $\text{Lat}(\text{LLM})/\text{Lat}(\text{Method})$.
}

\resizebox{\linewidth}{!}{
\begin{tabular}{l|c|ccc|ccc|ccc|ccc|ccc}
\toprule
\multirow{2}{*}{\textbf{Method}} & \multirow{2}{*}{$\boldsymbol{\tau}$} 
& \multicolumn{3}{c|}{AIME} 
& \multicolumn{3}{c|}{AMC} 
& \multicolumn{3}{c|}{Minerva} 
& \multicolumn{3}{c|}{MATH} 
& \multicolumn{3}{c}{OlympiadBench} \\
\cmidrule{3-17}
& & Acc $\uparrow$ & Lat. $\downarrow$ & Spd. $\uparrow$
& Acc $\uparrow$ & Lat. $\downarrow$ & Spd. $\uparrow$
& Acc $\uparrow$ & Lat. $\downarrow$ & Spd. $\uparrow$
& Acc $\uparrow$ & Lat. $\downarrow$ & Spd. $\uparrow$
& Acc $\uparrow$ & Lat. $\downarrow$ & Spd. $\uparrow$ \\

\midrule
\multicolumn{17}{c}{\textbf{Decoding budget = 8k tokens}} \\
\midrule
\rowcolor{gray!10}
SLM         & --   &  10.00   &  5.91   &  --   &  50.60   &  5.37   &  --   &  25.37   &   5.03  &  --  &  73.60  &  4.56   &  --   &  36.89  &  5.42  &  --   \\
\rowcolor{gray!15}
LLM-7B      & --   & 33.33 & 86.63 & 1.00$\times$ & 66.27 & 63.15 & 1.00$\times$ & 31.62 & 41.79 & 1.00$\times$ & 86.00 & 38.34 & 1.00$\times$ &  51.85  &  117.31  & 1.00$\times$ \\
SpecDec     & --   & 36.67 & 201.23 & 0.43$\times$ & 69.88 & 95.42 & 0.66$\times$ & 34.19 & 56.59 & 0.74$\times$ & 87.00 & 48.71 & 0.79$\times$ & 51.85 & 134.23 & 0.87$\times$  \\
\methodname & 0.001 & 36.67 & 89.86 & 0.96$\times$ & 77.11 & 58.72 & 1.08$\times$ & 34.19 & 38.73 & 1.08$\times$ & 89.40 & 32.94 & 1.16$\times$ &  55.26  &  105.29  &  1.11$\times$   \\
\methodname & 0.02  & 40.00 & 62.03 & 1.40$\times$ & 69.88 & 34.89 & 1.81$\times$ & 33.09 & 18.98 & 2.20$\times$ & 87.00 & 16.61 & 2.31$\times$ &  51.85  &  70.43  &  1.67$\times$   \\
\methodname & 0.03  & 30.00 & 42.06 & 1.84$\times$ & 69.88 & 24.55 & 2.57$\times$ & 33.09 & 15.07 & 2.77$\times$ & 85.60 & 15.15 & 2.53$\times$ &  48.59  &  51.43  &  2.28$\times$   \\
R-Stitch$^+$& --    & 40.00 & 37.19 & 2.33$\times$ & 68.67 & 21.08 & 3.00$\times$ & 35.29 & 15.33 & 2.73$\times$ & 86.60 & 15.68 & 2.45$\times$ &  52.00  & 45.02 & 2.34$\times$  \\
\rowcolor{gray!15}
LLM-14B     & --   & 43.33 & 153.20 & 1.00$\times$ & 68.67  &  101.76  & 1.00$\times$ & 35.29  & 48.02 & 1.00$\times$ &  86.00  &  41.66   & 1.00$\times$ & 54.07  &  190.89  & 1.00$\times$ \\
SpecDec     & --   & 50.00  & 139.10  & 1.10$\times$ & 68.67  &  95.59  &  1.06$\times$  &  34.93  & 53.02    &  0.91$\times$  &  82.80  &  45.88    &  0.91$\times$ &  54.22 & 180.03  &  1.06$\times$  \\
\methodname & 0.001 & 50.00  & 129.88  &  1.18$\times$  &  69.88  &  79.54  &  1.28$\times$  &  35.66  &  40.54  &  1.18$\times$  &  89.00  &  37.03  &  1.13$\times$ &  54.22  &  165.07  &  1.16$\times$   \\
\methodname & 0.02  & 43.33  &  87.61  &  1.75$\times$  &  69.88  &  41.82  &  2.43$\times$  &  35.66  &  22.16  &  2.17$\times$  &  85.20  &  20.53  &  2.05$\times$ &  52.44  &  96.09  &  1.99$\times$   \\
\methodname & 0.03  & 43.33  &  61.59  &  2.49$\times$  &  68.67  &  26.43  &  3.85$\times$  &  34.93  &  17.59  &  2.73$\times$  &  83.40  &  14.77  &  2.82$\times$ &  48.44  &  52.36  &  3.65$\times$   \\
R-Stitch$^+$& --    &  50.00  &  107.19  & 1.43$\times$ & 73.49  & 49.91   &  2.04$\times$    &  35.66   &  26.66  &  1.80$\times$   &  88.20  &  29.08  &  1.43$\times$  &  56.30  & 108.67  &  1.76$\times$   \\
\rowcolor{gray!15}
LLM-32B     & --   & 43.33  &  354.85  & 1.00$\times$ &   60.24  &  292.78  & 1.00$\times$ &  41.18  &  231.10  & 1.00$\times$ &  87.53  &  178.38  & 1.00$\times$ &  50.67  &  379.52  & 1.00$\times$ \\
SpecDec     & --   &  40.00  &  270.25  &  1.31$\times$  &  59.04  &  209.72  &  1.40$\times$  &  41.91  &  182.28  &  1.27$\times$  &  88.60  &  136.73   &  1.30$\times$  &  50.04  & 351.27   &  1.08$\times$  \\
\methodname & 0.001 & 50.00  &  261.86  &  1.40$\times$  &  68.75   &  209.88  &  1.39$\times$  &   42.65  &  141.24  &  1.64$\times$   &  91.20  &  95.49  &   1.87$\times$  &  50.67  &  341.67  &  1.11$\times$   \\
\methodname & 0.02  & 50.00  &  184.97  &  1.92$\times$  &   68.67  &  118.26  &  2.48$\times$  &  36.76   &  59.58  &  3.87$\times$  & 89.80  & 43.49  &  4.10$\times$  &  53.78  &  226.71  &  1.67$\times$   \\
\methodname & 0.03  & 40.00  &  178.19  &  1.99$\times$   &  69.88  &  86.88  &  3.37$\times$  &  34.56  &  43.41  &  5.32$\times$   &  87.00  &  32.21  &  5.54$\times$   &  52.15  &  157.47  & 2.41$\times$   \\
\midrule
\multicolumn{17}{c}{\textbf{Decoding budget = 16k tokens}} \\
\midrule
\rowcolor{gray!10}
SLM         & --   &  10.00   &  5.91   &  --   &  50.60   &  5.37   &  --   &  25.37   &   5.03  &  --  &  73.60  & 4.56   &  --   &  36.89  &  5.42  &  --   \\
\rowcolor{gray!15}
LLM-7B      & --   & 40.00 & 195.77 & 1.00$\times$ & 71.08 & 116.41    & 1.00$\times$ & 34.93 & 54.43 & 1.00$\times$ & 90.80 & 50.07 & 1.00$\times$ & 58.67 & 208.77 & 1.00$\times$ \\
SpecDec     & --   &   50.00  &   258.54  &  0.76$\times$  &  80.72   &  132.45   &  0.88$\times$  &  35.66  &  115.76  & 0.47$\times$ & 91.20  &  44.47     &  1.13$\times$  & 60.15 &  267.03  & 0.78$\times$    \\
\methodname & 0.001 & 46.67  &  192.22  &  1.03$\times$  & 80.72 &  97.26  & 1.20$\times$ & 35.66 &  45.53  & 1.20$\times$    &  91.00   &  38.34  & 1.31$\times$ & 60.15 &  201.13  & 1.04$\times$    \\
\methodname & 0.02  & 50.00 &  110.16  &  1.78$\times$  &  67.47  &  54.17  &  2.15$\times$   &  33.09  &  18.79  &  2.90$\times$  &  88.60   &  22.56  & 2.22$\times$ & 58.67 & 109.98  &  1.90$\times$   \\
\methodname & 0.03  & 36.67  &  41.51  &  4.72$\times$   &   66.27  &  38.07  &  3.06$\times$  &  35.29  &  13.50  &  4.03$\times$  &  83.20  & 15.62 & 3.21$\times$ & 54.22 &  65.65  & 3.18$\times$   \\
R-Stitch$^+$& --    &  50.00   &  86.03  &  2.28$\times$  &  77.11  &  56.03  &  2.08$\times$  &  35.29  &  13.68  &  3.98$\times$  &  88.60  &  16.02  &  3.13$\times$  &  59.11  & 95.31   &  2.19$\times$   \\
\rowcolor{gray!15}
LLM-14B     & --   &  50.00  &  246.95  & 1.00$\times$ &  86.75  &  146.51  & 1.00$\times$ &  39.34  &  68.99  & 1.00$\times$ &  88.60  &  66.54  & 1.00$\times$ & 60.00 &  316.06  & 1.00$\times$ \\
SpecDec     & --   &  50.00  &  275.64  & 0.90$\times$ &   83.13  & 138.53    &  1.06$\times$  &   39.34  &  55.40  &  1.25$\times$   &  87.40  &  67.18  & 0.99$\times$ & 60.00 & 360.14 & 1.05$\times$   \\
\methodname & 0.001 & 56.67  &  217.06  & 1.14$\times$ &  79.52  &  90.72   & 1.61$\times$ &  37.87  &  40.46  &  1.70$\times$  &  89.40   &   39.82  &  1.67$\times$ &  61.06 &  290.20  &  1.09$\times$   \\
\methodname & 0.02  & 43.33  &  99.12  &  2.49$\times$  &  66.27  &  51.29  &  2.86$\times$  &  31.99  &  22.74  &  3.04$\times$  &  87.80   &  22.62  &  2.94$\times$ & 54.22 &  108.61 &  2.91$\times$   \\
\methodname & 0.03  & 40.00  &  54.89  &  4.50$\times$  &  63.86  &  28.38  &  5.16$\times$  &  33.82  &  16.14  &  4.27$\times$  &  84.80   &  17.37  &  3.83$\times$ & 48.59 &  68.63 &  4.61$\times$   \\
R-Stitch$^+$& --    &  53.33  &  132.98  &  1.86$\times$  &  79.52  & 68.50    &  2.14$\times$  &  37.13  &  35.11  & 1.96$\times$ & 89.60  &  33.70  & 1.97$\times$ & 59.11 & 121.39  &  2.60$\times$   \\
\rowcolor{gray!15}
LLM-32B     & --   &   56.67  &  591.67  & 1.00$\times$ &  87.95  &  419.94  & 1.00$\times$ &  46.69  &  348.48  & 1.00$\times$ &  94.00  &  249.92  & 1.00$\times$ & 67.70 & 722.77 & 1.00$\times$ \\
SpecDec     & --   &  70.00  &  526.03  &  1.12$\times$  &  87.95  &  326.73   &  1.29$\times$  &  44.23  &  273.07  &  1.28$\times$  &  93.20  &  187.79  &  1.33$\times$   & 67.70 &  798.23  &  0.90$\times$   \\
\methodname & 0.001  &  70.00  &  488.87  &  1.21$\times$   &  90.36  &  288.92  &  1.45$\times$  &  41.18  &  126.62  &  2.75$\times$  &  94.00  &  172.08  &  1.45$\times$  & 66.07 & 654.24  & 1.10$\times$   \\
\methodname & 0.02  &  50.00  &  333.17  & 1.78$\times$   &  81.93   &  168.19  &  2.50$\times$  &  40.07  &  73.32  &  4.75$\times$  &  90.60  &  80.21  &  3.12$\times$  & 61.19  & 358.87  &  2.01$\times$   \\
\methodname & 0.03  &  53.33  &  276.03  & 2.14$\times$    &  73.49  &  143.53  &  2.93$\times$  &  36.40  &  43.58  &  8.00$\times$   &  87.80   &  38.32  &  6.52$\times$  & 55.70 & 209.96   &  3.44$\times$   \\
\bottomrule
\end{tabular}
}
\label{tab:math-hybrid-results}
\end{table*}

\textbf{Latency estimator.} However, directly measuring wall-clock latency during RL rollouts is impractical: trajectories are executed in batches and per-trajectory profiling would incur prohibitive overhead. To address this, we first sample a set of prefilling and decoding operations for each model (SLM and LLM) under different input lengths and KV cache size, and fit their latency profiles via linear regression. Let $N_{\text{inf}}$ denote the number of tokens processed at the current step and $N_{\text{kv}}$ the size of the KV cache before this step. The dominant cost arises from attention, which scales as $\mathcal{O}(N_{\text{inf}} \cdot N_{\text{kv}} + N_{\text{inf}}^2)$, plus linear terms in $N_{\text{inf}}$. Accordingly, we model the latency for a prefill operation as
\begin{equation}
\label{eq:prefill_time}
T_{\text{prefill}}(N_{\text{inf}}, N_{\text{kv}}) 
= a \cdot N_{\text{inf}} \cdot N_{\text{kv}} + b \cdot N_{\text{inf}}^2 + c \cdot N_{\text{inf}} + d,
\end{equation}
where coefficients $(a,b,c,d)$ are fitted via linear regression on profiling data. A decoding step corresponds to $N_{\text{inf}}=1$, yielding
\begin{equation}
\label{eq:decode_time}
T_{\text{decode}}(N_{\text{kv}}) = c \cdot N_{\text{kv}} + d.
\end{equation}
The trajectory-level latency $\widehat{L}$ is then obtained by summing the step-wise estimates. 

\textbf{DAPO optimization.}
We train the router with the DAPO optimizer~\citep{yu2025dapo}. 
For each prompt $(q,a)$, we sample a group of $G$ routed trajectories $\{o_i\}_{i=1}^G\!\sim\!\pi_{\theta_{\text{old}}}(\cdot\mid q)$ and compute a scalar reward $R_i$ for each trajectory.
Let $o_{i,t}$ be the $t$-th action in $o_i$.
We define the importance ratio and the group-normalized advantage as
\begin{equation}
\label{eq:dapo_ratio_adv}
\begin{aligned}
r_{i,t}(\theta)
&=
\frac{
  \pi_\theta\!\left(o_{i,t}\mid q,\,o_{i,<t}\right)
}{
  \pi_{\theta_{\text{old}}}\!\left(o_{i,t}\mid q,\,o_{i,<t}\right)
}, \\[0.5ex]
\hat{A}_{i,t}
&=
\frac{
  R_i - \operatorname{mean}\!\left(\{R_j\}_{j=1}^G\right)
}{
  \operatorname{std}\!\left(\{R_j\}_{j=1}^G\right)
}.
\end{aligned}
\end{equation}

Here, $\pi_\theta(\cdot)$ represents the current router policy, while $\pi_{\theta_{\text{old}}}(\cdot)$ denotes the policy used to sample trajectories in the previous iteration. Let $\mathrm{clip}_\varepsilon(x) = \min(\max(x,1-\varepsilon),\,1+\varepsilon)$ denote the clipping operator. The DAPO objective is
\begin{equation}
\label{eq:dapo}
\begin{aligned}
\mathcal{J}_{\text{DAPO}}(\theta)
&= \mathbb{E}_{(q,a),\,\{o_i\}\sim\pi_{\theta_{\text{old}}}}
\Bigg[
  \frac{1}{\sum_{i=1}^{G} |o_i|}
  \sum_{i=1}^{G}\sum_{t=1}^{|o_i|} \\
&\qquad\min\!\Big(
        r_{i,t}(\theta)\,\hat{A}_{i,t},\;
        \mathrm{clip}_\varepsilon\!\big(r_{i,t}(\theta)\big)\,\hat{A}_{i,t}
      \Big)
\Bigg]
\end{aligned}
\end{equation}

where $\varepsilon$ is the clipping threshold.

\begin{figure*}[t]
    \centering

    \begin{minipage}{0.9\linewidth}
        \centering
        \includegraphics[width=\linewidth]{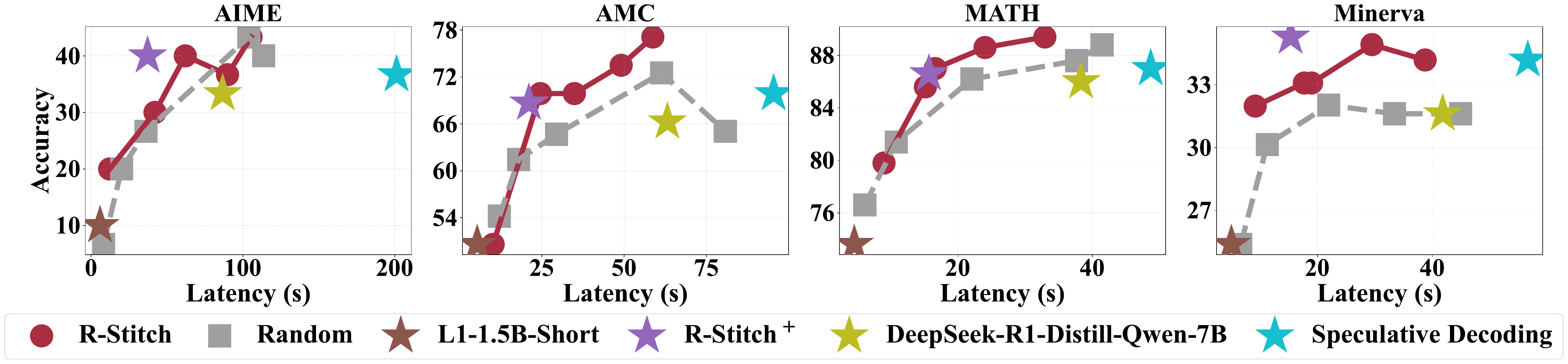}
    \end{minipage}

    \begin{minipage}{0.9\linewidth}
        \centering
        \includegraphics[width=\linewidth]{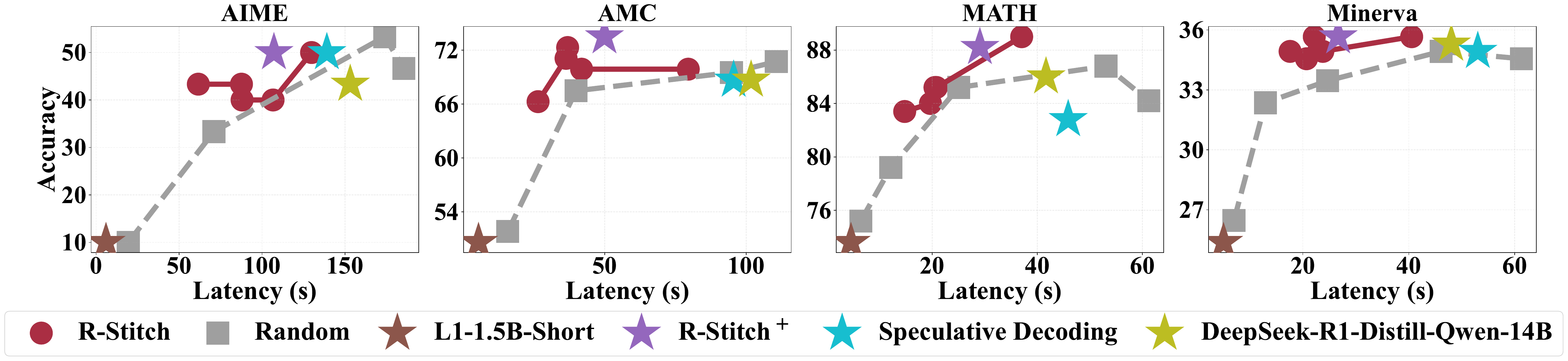}
    \end{minipage}
    \vspace{-1.5mm}
    \caption{
Accuracy–latency trade-off curves on mathematical reasoning datasets. 
The first row uses the 7B LLM, and the second row uses the 14B LLM. 
The red lines correspond to our method (\methodname) under varying entropy thresholds $\tau$ and a random routing baseline. 
}

    \label{fig:latency-accuracy-curve}
\end{figure*}



\section{Experiments}
\label{sec:experiments}

\noindent\textbf{Implementation details.}
We evaluate our proposed method on five mathematical reasoning benchmarks: OlympiadBench~\citep{he2024olympiadbench}, AIME~\citep{li2024numinamath}, Minerva~\citep{lewkowycz2022solving}, AMC~\citep{li2024numinamath}, and MATH~\citep{hendrycks2021measuring}.  We adopt DeepSeek-Math-R1-Distill-Qwen models at 7B and 14B scales, and QwQ-32B as the 32B-scale LLM, with L1-Short serving as the SLM. Our method is implemented within the vLLM~\citep{kwon2023vllm} inference framework, where the LLM and SLM are instantiated as separate engines, each maintaining its own key-value cache to enable independent decoding and efficient model switching. Speculative decoding baselines are evaluated using the official vLLM speculative decoding code.
For \methodname$^+$, we train the router using DAPO~\citep{yu2025dapo}’s publicly released RL training dataset, with $\lambda = 5 \times 10^{-6}$, $\tau = 0.001$,  batch size $32$, and rollout group size $8$. The latency estimator is calibrated in milliseconds. Due to GPU memory constraints, RL experiments are conducted only on the 7B and 14B models.
All latency numbers reported in the results correspond to the average wall-clock inference time per sample on a single NVIDIA A100 GPU with batch size 1.

\begin{figure*}[h]
    \centering
    \includegraphics[width=0.92\linewidth]{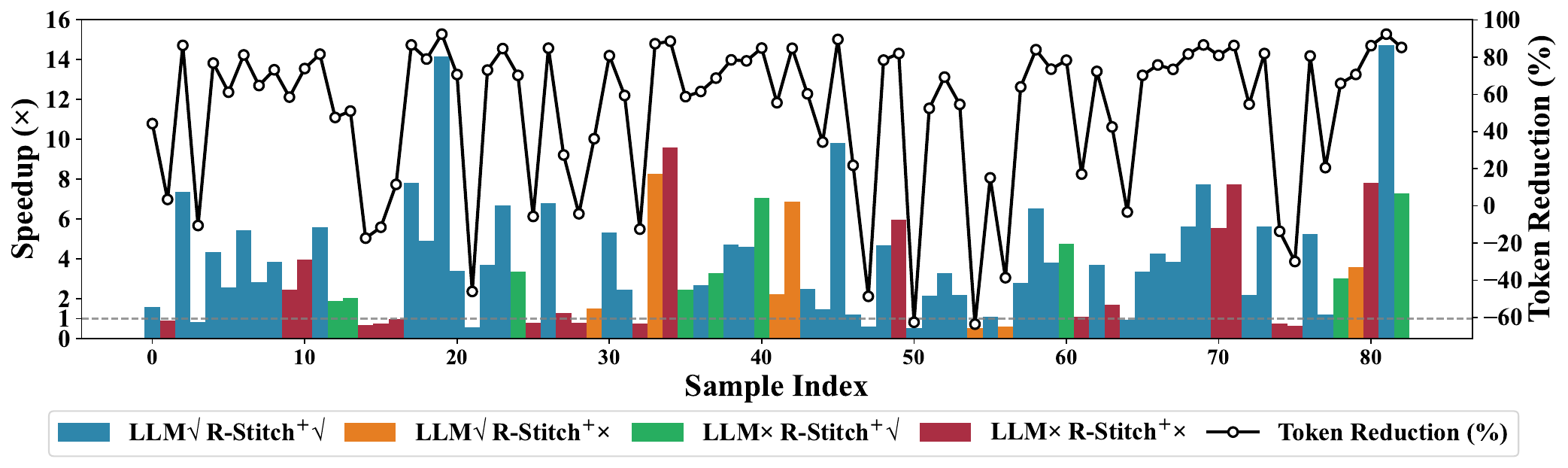}
    \includegraphics[width=0.92\linewidth]{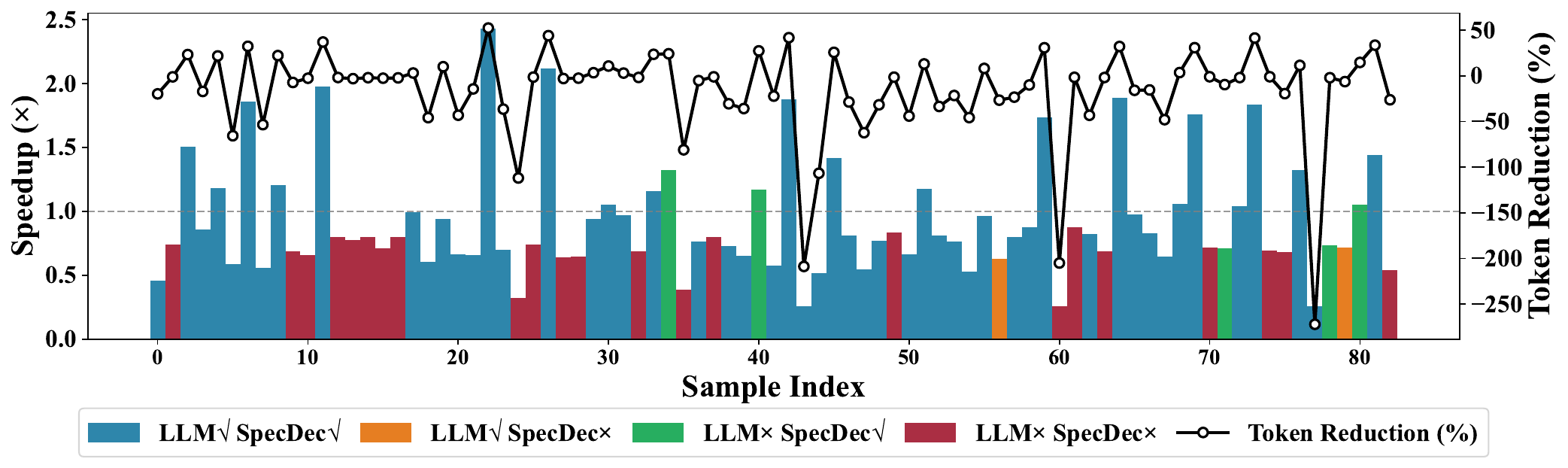}
    \vspace{-2.5mm}
    \caption{\textbf{Per-sample visualization of R-Stitch$^{+}$ and Speculative Decoding (LLM-7B).}  
    Each bar shows the latency speedup of one sample relative to the baseline LLM.  
    Bar colors encode correctness outcomes of the baseline and R-Stitch$^{+}$.  
    The dashed horizontal line at 1 indicates no speedup.  
    The black solid curve with hollow-circle markers represents token reduction percentages per sample.}
    \label{fig:rstitch-6-all}
\end{figure*}

\noindent\textbf{Main results on math reasoning benchmarks.}~~
We evaluate \methodname~under entropy thresholds $\tau \in \{0.001, 0.02, 0.03\}$, comparing against SLM-only decoding, full LLM decoding, and speculative decoding (SpecDec). 
Table~\ref{tab:math-hybrid-results} reports accuracy, latency, and relative speedup across five mathematical reasoning benchmarks under decoding budgets of 8k and 16k tokens. 
Overall, \methodname~consistently reduces latency with minimal accuracy degradation relative to full LLM decoding. At $\tau=0.02$, LLM-7B achieves $1.4\times$–$2.3\times$ speedups at 8k and $1.7\times$–$2.9\times$ at 16k, while maintaining accuracy close to the LLM baseline. For larger models, efficiency gains become more pronounced: LLM-14B reaches $1.7\times$–$2.4\times$ at 8k and up to $3.0\times$ at 16k, while LLM-32B attains speedups of $1.9\times$–$4.1\times$ at 8k and $1.8\times$–$4.8\times$ at 16k. \rebuttal{In particular, the observed accuracy gains arise because R-Stitch shortens overly verbose LLM reasoning traces, preventing truncation under strict length budgets and thereby avoiding accuracy degradation.} SpecDec shows mixed performance: while occasionally competitive, it often runs slower than standard LLM decoding (e.g., 7B at 8k) or sacrifices accuracy. 
This slowdown is largely due to frequent rejections under low model agreement, which cause the tokens of SLM to be discarded and re-decoded by the LLM, resulting in more total decoding steps than even vanilla LLM decoding. 
SLM-only decoding remains fastest but yields substantially lower accuracy, highlighting the necessity of hybrid approaches. These results demonstrate that entropy-based routing provides a robust trade-off between efficiency and accuracy across different model scales and token budgets, with larger models benefiting the most in relative speedups.

\noindent\textbf{Latency–accuracy trade-off.}~~
We further assess entropy-guided routing by varying the threshold $\tau \in \{0.001, 0.005, 0.02, 0.03, 0.05\}$ and reporting the resulting accuracy–latency trade-off. As baselines, we consider random routing, where tokens are assigned to the SLM with fixed probabilities $p \in \{0.1, 0.3, 0.5, 0.7, 0.9\}$, along with fixed decoding strategies including SLM-only, LLM-only, and speculative decoding. Figure~\ref{fig:latency-accuracy-curve} presents results on four mathematical reasoning benchmarks with both 7B and 14B LLMs. Across all settings, \methodname consistently outperforms random routing, confirming entropy as a reliable confidence signal. On relatively simple tasks such as AMC, it achieves accuracy comparable to full LLM decoding while operating at substantially lower latency. Overall, entropy-based routing offers a simple, training-free mechanism that effectively balances efficiency and accuracy across diverse datasets and model scales. In addition, by tuning the entropy threshold, \methodname naturally adapts to different computational budgets without retraining, enabling practitioners to approach optimal accuracy under varying latency constraints.

\noindent\textbf{Per-sample comparison with Speculative Decoding.} Figure \ref{fig:rstitch-6-all} provides per-sample visualizations of latency speedup and token reduction when applying R-Stitch$^{+}$ and Speculative Decoding on the LLM-7B in AMC. We observe that R-Stitch$^{+}$ consistently accelerates the majority of samples, while Speculative Decoding only yields speedup on a small portion of samples and causes slowdowns on most due to consistency issues. Notably, the theoretical upper bound of speculative decoding’s speedup under high consistency is the latency ratio between the LLM and SLM. In the visualization, the fastest sample achieves just above a $2\times$ speedup, which matches the decoding latency gap between the 7B and 1.5B models. \textit{In contrast, our method additionally leverages the concise expressiveness of the SLM while preserving accuracy, thereby achieving substantially higher acceleration of up to \textbf{14$\times$}.}

\begin{table}[h]
\centering
\small
\caption{Memory usage of SLM/LLM models and their KV caches under BF16.
\textbf{KV/tok} denotes per-token KV cache size in MB, \textbf{Wts} denotes model weights in MB,
and \textbf{Total@8k}/\textbf{Total@16k} report total memory usage at 8k/16k context lengths.}
\begin{tabular}{lcccc}
\toprule
\textbf{Model} & \textbf{KV/tok} & \textbf{Wts} & \textbf{Total@8k} & \textbf{Total@16k} \\
\midrule
1.5B  & 0.028 & 3036  & 3265  & 3494  \\
7B    & 0.055 & 15640 & 16091 & 16542 \\
14B   & 0.189 & 29275 & 30824 & 32374 \\
32B   & 0.253 & 63890 & 65958 & 68026 \\
\bottomrule
\end{tabular}
\label{tab:kv_memory}
\end{table}

\textbf{Memory Overhead of Dual KV Caches.}~R-Stitch, like all speculative-decoding–style approaches, maintains separate KV caches for the SLM and LLM. To quantify the actual overhead, we measure model weights and KV footprints in \texttt{vLLM} under BF16 precision. Table~\ref{tab:kv_memory} reports per-token KV size, model weights, and total memory usage at 8k and 16k context lengths. When paired with a 7B model under a 16k context, the 1.5B SLM introduces only a 17.44\% memory increase, which further decreases to 9.74\% for a 14B model and 4.89\% for a 32B model. These results show that the additional memory required by R-Stitch remains modest relative to the LLM footprint, while enabling substantial end-to-end latency reductions.

\section{Conclusion}
This paper has introduced \methodname, a dynamic routing strategy for accelerating large language model inference by selectively delegating token-level computation to a SLM. Our method leverages an entropy-based switching mechanism that routes easy tokens to the lightweight model while preserving overall output quality. This design enables a favorable trade-off between inference latency and accuracy without requiring additional retraining or architectural changes. 
Furthermore, we have extended this approach to \methodname$^+$, which incorporates a reinforcement learning (RL)–based router trained with a latency-aware reward. By adaptively deciding when to switch between the SLM and the LLM, \methodname$^+$ refines the efficiency–accuracy balance beyond fixed entropy thresholds, achieving stronger acceleration while maintaining high accuracy. Comprehensive experiments on multiple datasets have demonstrated that our framework achieves consistent improvements over static baselines and speculative decoding, offering a practical and efficient solution for real-world deployment of LLMs.

\textbf{Limitations and future work.}~Our current implementation supports only batch size 1 due to dynamic token-level model switching, which restricts hardware utilization in practical deployment. Addressing this limitation may require designing new scheduling strategies or restructuring the routing mechanism to better accommodate batched inference. 
To further alleviate the KV cache and parameter burden from maintaining two models, we plan to adopt parameter-sharing strategies to enhance memory efficiency. These extensions could improve the robustness and scalability of dynamic model routing, and enable more fine-grained control over the latency–accuracy trade-off.

\section*{Impact Statement}

This paper presents work aimed at advancing the field of machine learning by improving the efficiency of inference-time reasoning in large language models. The proposed method reduces computational cost and latency while maintaining accuracy, which may facilitate more efficient and accessible deployment of reasoning-capable models.
The societal and ethical implications of this work are consistent with those commonly associated with improvements in machine learning efficiency, including potential benefits in sustainability and practical usability. The method operates purely at inference time and does not introduce new model capabilities, data sources, or application domains, and therefore does not raise new ethical concerns beyond those already well established for large language models.
Overall, we do not identify any broader impacts that require specific discussion beyond these considerations.

\bibliography{main}
\bibliographystyle{icml2026}

\newpage
\appendix
\onecolumn
\section{Appendix}



\subsection{Inference algorithm of the proposed method}

\begin{algorithm}[h]
\caption{Inference procedure of the proposed method}
\label{alg:r-stitch-infer}
\small
\begin{algorithmic}[1]
\REQUIRE Prompt $x$; models \textsc{SLM}, \textsc{LLM}; threshold $\tau$; max length $T_{\max}$
\STATE Optional: router $\pi_\theta$ (\methodname$^+$)

\STATE $\mathsf{KV}^{\textsc{SLM}}\!\gets\emptyset,\ \mathsf{KV}^{\textsc{LLM}}\!\gets\emptyset$; $L^{\textsc{SLM}}\!=L^{\textsc{LLM}}\!=0$ \; $\triangleright$ $\mathsf{KV}^{\textsc{SLM}}$, $\mathsf{KV}^{\textsc{LLM}}$: KV caches; $L^{\textsc{SLM}}$, $L^{\textsc{LLM}}$: KV cache sizes of $\textsc{SLM}$ and $\textsc{LLM}$
\STATE $y\gets[]$, $\textsc{active}\gets\textsc{SLM}$ \; $\triangleright$ $y$: generated output; $\textsc{active}$: current inference model

\FOR{$t=1$ $\to$ $T_{\max}$}
    \IF{$L^{\textsc{active}} < |x\oplus y|$} 
    \STATE \textbf{Prefill}: run \textsc{active} on $(x\oplus y)[L^{\textsc{active}}{:}]$ to update cache and generate $(\mathbf{p}_t,y_t)$; $L^{\textsc{active}}\gets|x\oplus y|+1$ 
    \; $\triangleright$ KV cache not up-to-date; $\mathbf{p}_t$: prob. distribution; $y_t$: generated token
\ELSE
    \STATE \textbf{Decode}: run \textsc{active}$\,$ with cache to generate $(\mathbf{p}_t,y_t)$; $L^{\textsc{active}}\gets L^{\textsc{active}}+1$
\ENDIF

    \STATE Compute entropy $\mathcal{H}_t$ from $\mathbf{p}_t$ (Eq.~\ref{eq:entropy}) \; $\triangleright$ $\mathcal{H}_t$: token entropy
    \IF{$\mathcal{H}_t > \tau$}
        \IF{\methodname}
            \STATE $\textsc{switch}\gets$ True  
        \ELSIF{\methodname$^+$}
            \STATE $\textsc{switch}\gets (\arg\max_{a\in\{\textsc{SLM},\textsc{LLM}\}}\pi_\theta(a\mid s_t)=\textsc{LLM})$ \; $\triangleright$ router decision (state $s_t$)
        \ENDIF
        \IF{$\textsc{active}=\textsc{SLM}$ \textbf{and} $\textsc{switch}$}
            \STATE \textbf{Discard} $y_t$ and rollback $\mathsf{KV}^{\textsc{SLM}}$; $\textsc{active}\gets\textsc{LLM}$; \textbf{continue} \; $\triangleright$ switch SLM$\to$LLM, token not kept
        \ENDIF
    \ELSIF{$\textsc{active}=\textsc{LLM}$}
        \STATE $\textsc{active}\gets\textsc{SLM}$ \; $\triangleright$ switch LLM$\to$SLM
    \ENDIF

    \STATE Append $y_t$ to $y$; update cache \; $\triangleright$ append only if token not discarded
    \IF{$y_t=\texttt{[EOS]}$} \STATE \textbf{break} \ENDIF
\ENDFOR

\STATE \textbf{return} $y$
\end{algorithmic}
\end{algorithm}

\subsection{Latency Regression Results}
\label{app:latency-regression}

Building on the latency estimator introduced in Eq.~\ref{eq:prefill_time} and Eq.~\ref{eq:decode_time}, 
we report the fitted coefficients $(a, b, c, d)$ obtained from profiling three representative models: SLM-1.5B, LLM-7B, and LLM-14B. 
Recall that $N_{\text{inf}}$ denotes the number of tokens processed at the current step and $N_{\text{kv}}$ the size of the KV cache before this step. 
Latency is modeled as 
\[
T(N_{\text{inf}}, N_{\text{kv}}) = a \cdot N_{\text{inf}} \cdot N_{\text{kv}} + b \cdot N_{\text{inf}}^2 + c \cdot N_{\text{inf}} + d.
\]

The fitted functions are as follows (latency $T$ in milliseconds):

\begin{itemize}  
  \item SLM-1.5B:
  \[
  T(N_{\text{inf}}, N_{\text{kv}}) = 0.000021 \cdot (N_{\text{inf}} \cdot N_{\text{kv}})
    + 0.000231 \cdot N_{\text{inf}}^2 
    - 0.121046 \cdot N_{\text{inf}} 
    + 27.090929
  \]

  \item LLM-7B:
  \[
  T(N_{\text{inf}}, N_{\text{kv}}) = 0.000027 \cdot (N_{\text{inf}} \cdot N_{\text{kv}})
    + 0.000031 \cdot N_{\text{inf}}^2 
    - 0.045256 \cdot N_{\text{inf}} 
    + 27.040801
  \]

  \item LLM-14B:
  \[
  T(N_{\text{inf}}, N_{\text{kv}}) = 0.000045 \cdot (N_{\text{inf}} \cdot N_{\text{kv}})
    + 0.000123 \cdot N_{\text{inf}}^2 
    - 0.082998 \cdot N_{\text{inf}} 
    + 45.118931
  \]
\end{itemize}

These regression results provide practical estimators for per-step latency under varying input lengths and KV cache sizes, enabling efficient evaluation without costly profiling.

\subsection{Related work of optimizing Speculative decoding}

Recent approaches to speculative decoding acceleration can be understood as attempts to 
train auxiliary mechanisms that increase the consistency between the draft model and the target model. 
This line of research is exemplified by the EAGLE family and Griffin, while Hydra represents a 
lighter-weight but still consistency-oriented variant, and CITER departs somewhat from this trend 
but nevertheless introduces additional training cost. We briefly review these methods below. 

\textbf{EAGLE.} 
The EAGLE series~\citep{li2024eagle,li2024eagle2,li2025eagle3} develops progressively more sophisticated 
training strategies for draft models. 
The original EAGLE~\citep{li2024eagle} proposed to autoregressively predict the top-layer hidden features 
of the target model, rather than tokens directly, thereby anchoring the draft model’s outputs more closely 
to the target distribution. This was combined with a tree-based drafting mechanism and corresponding 
tree attention to improve parallelism. 
EAGLE-2~\citep{li2024eagle2} extended this by introducing a dynamic draft tree that adapts its expansion 
based on the draft model’s confidence, reducing wasted computation on unlikely branches. 
Finally, EAGLE-3~\citep{li2025eagle3} removed the restriction of feature-level prediction and returned to token-level generation, 
but enhanced it with multi-level feature fusion and a novel training-time test technique, 
which simulates self-feedback during training. These changes allow EAGLE-3 to scale better with more training data 
and to reach higher acceptance rates, but all variants require substantial supervised training of the draft model.

\textbf{Griffin.} 
Griffin~\citep{hu2025griffin} addresses the bottleneck of speculative decoding by improving the quality of the small draft model rather than modifying the large model or the verification process. 
It introduces guidance distillation to better align the draft model’s predictive distribution with that of the target model, and augments the architecture with lightweight enhancements that increase expressiveness without incurring significant cost. 
As a result, Griffin achieves substantially higher acceptance rates than naive draft models, thereby unlocking more consistent acceleration for speculative decoding.

\textbf{Hydra.} 
Hydra~\citep{ankner2024hydra} departs from training a separate small model and instead equips the large model 
with a set of Hydra heads, lightweight proposal modules attached to intermediate hidden states. 
Unlike the earlier Medusa framework~\citep{cai2024medusa}, where draft heads were independent and predicted tokens in parallel, 
Hydra enforces sequential dependency across heads: each head conditions on the tokens predicted by previous heads. 
This design increases the internal consistency of the drafted sequence, leading to higher acceptance rates 
without the need for an external small model. 
Hydra thus offers a lighter-weight solution, but it still shares the central assumption that 
greater draft–target consistency is the key to efficiency.

\textbf{Distinctive perspective of our method.} 
In contrast to all these approaches, our work takes a fundamentally different perspective. 
We do not attempt to make the small and large models more consistent, nor do we introduce 
training overhead for rewriting. Instead, we explicitly exploit the insight that the small model is 
\emph{simpler and naturally inconsistent} with the large model. 
Rather than mitigating this inconsistency, we turn it into an advantage, showing that it can be directly 
leveraged to improve speculative decoding efficiency. 
Training-free R-Stitch requires \emph{no additional training} and no auxiliary modules, 
yet achieves significant acceleration. 
This reveals a new perspective: sometimes inconsistency between models is not a limitation to be avoided, 
but a property that can be actively harnessed for efficiency gains.

\begin{table}[h]
\centering
\caption{
Performance comparison with the state-of-the-art speculative decoding baseline (Eagle-3, the fastest method reported in the official vLLM repository) on Qwen3-8/14B~\citep{yang2025qwen3} across four mathematical reasoning benchmarks. 
We report accuracy (Acc), latency (s/sample), average output length (Tok.), and relative speedup (Spd.), under both 8k-token and 16k-token decoding budgets.
}
\label{tab:eagle}
\resizebox{\linewidth}{!}{
\begin{tabular}{l|c|cccc|cccc|cccc|cccc}
\toprule
\multirow{2}{*}{\textbf{Method}} & \multirow{2}{*}{$\boldsymbol{\tau}$} 
& \multicolumn{4}{c|}{AIME} 
& \multicolumn{4}{c|}{AMC} 
& \multicolumn{4}{c|}{Minerva} 
& \multicolumn{4}{c}{MATH} \\
\cmidrule{3-18}
& & Acc $\uparrow$ & Lat. $\downarrow$ & Tok. $\downarrow$ & Spd. $\uparrow$
  & Acc $\uparrow$ & Lat. $\downarrow$ & Tok. $\downarrow$ & Spd. $\uparrow$
  & Acc $\uparrow$ & Lat. $\downarrow$ & Tok. $\downarrow$ & Spd. $\uparrow$
  & Acc $\uparrow$ & Lat. $\downarrow$ & Tok. $\downarrow$ & Spd. $\uparrow$ \\
\midrule
\multicolumn{18}{c}{\textbf{Decoding budget = 8k tokens}} \\
\midrule
\rowcolor{gray!15}
Qwen3-8B      & --   & 30.00 & 97.50 & 7589.33 & 1.00$\times$ & 59.04 & 81.68 & 6384.39  & 1.00$\times$ & 34.93 & 68.49 & 5367.97  & 1.00$\times$ & 84.00  & 56.02 & 4423.09  & 1.00$\times$ \\
Eagle-3 & --   & 36.67 & 64.87 & 7740.33 & 1.50$\times$ & 59.04 & 52.35 & 6575.19 & 1.56$\times$  & 36.03 & 42.30  & 5380.57 & 1.62$\times$  & 86.00 & 33.87 & 4437.59 & 1.65$\times$ \\
\methodname & 0.001 & 50.00 & 64.89 & 6978.97 & 1.50$\times$ & 71.08 & 44.72 & 4942.05 & 1.83$\times$ & 42.65 & 35.68 & 3913.13 & 1.92$\times$ & 89.60 & 24.77 & 2845.66 &  2.26$\times$  \\
\methodname & 0.01  & 56.67 & 48.30 & 5527.50 & 2.02$\times$ & 62.65 & 27.53 & 3316.34 & 2.97$\times$ & 38.97 & 27.53 & 2030.41 & 2.49$\times$ & 87.00 & 12.52 & 1614.31 & 4.47$\times$  \\
\methodname & 0.02  & 26.67 & 39.35 & 4701.90 & 2.48$\times$ & 66.27 & 19.27 & 2453.30 & 2.97$\times$  & 34.56 & 10.86 & 1416.57 & 6.31$\times$  & 83.60 & 9.27 & 1260.76  & 6.04$\times$  \\
\rowcolor{gray!15}
Qwen3-14B      & --   & 40.00 & 158.80 & 7549.47 & 1.00$\times$ & 62.65 & 126.86 & 6055.57 & 1.00$\times$ & 41.54 & 100.52 & 4808.01 & 1.00$\times$ & 89.00 & 82.93 & 3980.87 & 1.00$\times$ \\
Eagle-3     & --   & 43.33 & 110.48 & 7700.53 & 1.44$\times$ & 60.24 & 83.70 & 6125.43 & 1.52$\times$ & 39.71 & 67.10 & 4891.85  & 1.50$\times$ & 88.20 & 53.89 & 4056.89 & 1.54$\times$  \\
\methodname & 0.001 & 46.67 & 94.97 & 7043.47 & 1.67$\times$ & 67.47 & 63.13 & 4802.56 & 2.01$\times$ &  45.59 & 46.30 & 3545.74 & 2.17$\times$ & 90.00 & 32.62 & 2688.23 & 2.54$\times$  \\
\methodname & 0.01  & 43.33 & 75.51 & 6213.47 &  2.10$\times$ & 65.06 & 35.75 & 3213.11 & 3.55$\times$ & 40.44 & 20.53 & 1815.11 & 4.90$\times$ & 88.00 & 16.24 & 1569.15 & 5.11$\times$  \\
\methodname & 0.02  & 33.33 & 55.20 & 4836.30 & 2.88$\times$ & 61.45  & 23.99 & 2330.29 & 5.29$\times$ &  37.13 & 13.01 & 1266.13  & 7.73$\times$ & 84.60 & 11.64 & 1223.35 & 7.12$\times$  \\
\midrule
\multicolumn{18}{c}{\textbf{Decoding budget = 16k tokens}} \\
\midrule
\rowcolor{gray!15}
Qwen3-8B      & --   & 63.33 & 156.21 & 11899.80  & 1.00$\times$ & 81.93 & 115.33 & 9022.71  & 1.00$\times$ & 47.43 & 90.28 & 6822.03  & 1.00$\times$ & 94.40 & 65.05 & 5067.67 & 1.00$\times$ \\
Eagle-3       & --   & 66.67 & 110.23 & 12080.73 & 1.42$\times$ & 81.93 & 76.55 & 9008.47 & 1.51$\times$ & 47.06 & 58.49 & 6981.51 & 1.54$\times$ & 95.60 & 40.60 & 5150.57 & 1.60$\times$  \\
\methodname   & 0.0005 & 70.00 & 109.82 & 10942.50  & 1.42$\times$ & 81.93 & 70.46 & 7107.68 & 1.64$\times$ & 46.32  & 48.65 & 3640.69 & 1.86$\times$ & 94.60 & 33.64 & 3640.69 & 1.93$\times$  \\
\methodname   & 0.001  & 60.00 & 102.83 & 10135.03 & 1.52$\times$ &  74.70 & 65.14 & 6606.36 & 1.77$\times$ & 41.54 & 44.73 & 4510.72 & 2.02$\times$ &  94.80 & 29.20 & 3238.44 & 2.23$\times$  \\
\methodname   & 0.01  & 53.33 & 69.63 & 7506.93 & 2.24$\times$ & 73.49 & 38.13 & 4319.67 & 3.02$\times$ & 37.50  & 18.04 & 2142.44 & 5.00$\times$ & 89.60  & 14.97 & 1839.84 & 4.35$\times$  \\
\rowcolor{gray!15}
Qwen3-14B     & --   & 70.00 & 248.17 & 11237.64 & 1.00$\times$ & 81.93 & 167.32 & 7896.86  & 1.00$\times$ & 46.69 & 125.74 & 5959.92  & 1.00$\times$ & 95.60 & 94.76 & 4528.12  & 1.00$\times$ \\
Eagle-3       & --   & 63.33 & 189.59 & 11203.67 & 1.31$\times$ & 85.54 & 121.30 & 8292.41 & 1.38$\times$ & 47.06 & 84.50 & 5930.95 & 1.49$\times$ & 95.40 & 63.31 & 4616.20 & 1.50$\times$   \\
\methodname   & 0.0005 & 70.00 & 139.11 & 9637.84  & 1.78$\times$ & 85.54 & 89.25 & 6346.75 & 1.87$\times$ &  45.96 & 59.48 & 4373.63 & 2.11$\times$ & 94.60 & 45.60 & 3427.84 & 2.08$\times$  \\
\methodname   & 0.001  & 70.00 & 136.51 & 9556.70  & 1.82$\times$ & 83.13 & 91.11 & 6713.41 & 1.84$\times$ & 44.85 & 53.10 & 3985.24 & 2.37$\times$ & 94.20 & 37.34 & 3003.98 & 2.54$\times$  \\
\methodname   & 0.01  & 56.67 & 113.50 & 8668.10 & 2.19$\times$ & 71.08 & 40.84 & 3512.52 & 4.10$\times$ &  38.24 & 24.58 & 2116.48 & 5.12$\times$ & 88.60 & 20.65 & 1868.24 & 4.59$\times$  \\
\bottomrule
\end{tabular}
}
\end{table}

\subsection{Comparison with State-of-the-Art Speculative Decoding}
\label{sec:appendix-sota-specdec}

Table~\ref{tab:eagle} compares our method with Eagle-3, the state-of-the-art speculative decoding approach reported in the official vLLM repository, on Qwen3-8B and Qwen3-14B across four mathematical reasoning benchmarks under both 8k and 16k decoding budgets. Both Eagle-3 and R-Stitch exploit the advantage that small models decode tokens faster than the LLM, thereby reducing per-token complexity. However, R-Stitch additionally leverages the conciseness of SLM outputs: by selectively retaining short, low-entropy continuations from the SLM, it reduces the overall output length, yielding substantially higher end-to-end acceleration. Notably, Eagle-3 employs a heavily trained draft model much smaller than our 1.5B SLM, making its per-token decoding inherently faster. Despite this disadvantage, R-Stitch achieves consistently larger speedups across all benchmarks and budgets, highlighting the benefit of flexible entropy-guided collaboration over purely consistency-driven speculative decoding.

\subsection{Combining with Early Exit Strategies}
\begin{table}[h]
\centering
\caption{
Effect of applying R-Stitch ($\tau=0.01$) on top of DEER with the LLM-7B under a 16k-token decoding budget. 
Accuracy (\%), average token count, and inference latency (s) are reported. 
Results of the vanilla LLM-7B are also included for reference.
}
\label{tab:deer_rstitch_split}
\resizebox{0.95\linewidth}{!}{
\begin{tabular}{l ccc ccc ccc}
\toprule
\multirow{2}{*}{\textbf{Dataset}} 
& \multicolumn{3}{c}{\textbf{LLM-7B}} 
& \multicolumn{3}{c}{\textbf{+ DEER}} 
& \multicolumn{3}{c}{\textbf{+ DEER + R-Stitch}} \\
\cmidrule(lr){2-4}\cmidrule(lr){5-7}\cmidrule(lr){8-10}
& Acc. (\%) & Token & Lat. (s) 
& Acc. (\%) & Token & Lat. (s) 
& Acc. (\%) & Token & Lat. (s) \\
\midrule
MATH   & 90.80 & 3381.51 & 50.07 & 89.20 & 2284.52 & 35.47 & 88.40 & 1564.90 & 20.59 \\
AIME   & 40.00 & 11739.10 & 195.77 & 36.67 & 9229.03 & 210.37 & 36.67 & 4425.20 & 92.52 \\
GSM8K  & 88.93 & 1398.54 & 20.91 & 90.07 & 698.47 & 9.69 & 90.07 & 462.41 & 6.76 \\
GPQA-D & 23.74 & 10134.81 & 171.76 & 27.78 & 7084.15 & 117.50 & 30.81 & 1047.85 & 18.15 \\
\bottomrule
\end{tabular}}
\end{table}

We further examine whether \methodname~can complement training-free early-exit methods to improve decoding efficiency. 
Specifically, we combine \methodname~with DEER~\citep{yang2025dynamic}, which halts generation once the model’s confidence surpasses a predefined threshold. 
As shown in Table~\ref{tab:deer_rstitch_split}, evaluated on MATH~\citep{hendrycks2021measuring}, AIME~\citep{li2024numinamath}, GSM8K~\citep{cobbe2021training}, and GPQA-D~\citep{rein2024gpqa} under a 16k-token budget, the hybrid system substantially reduces both token count and end-to-end latency compared to DEER alone, while maintaining comparable accuracy. On AIME, token usage is reduced by more than 50\% and latency drops from 210.37s to 92.52s, without affecting accuracy. 
On GPQA-D, the benefits are even more pronounced: latency decreases from 117.50s to 18.15s, while accuracy improves from 27.78\% to 30.81\%. Compared to vanilla LLM-7B decoding (171.76s), this corresponds to a \textbf{$9.5\times$} acceleration.
These results confirm that early exiting and entropy-guided routing are complementary: DEER shortens the reasoning trajectory, whereas \methodname~lowers per-token cost by selectively invoking the LLM only when needed. 
Together, they provide strictly greater efficiency by simultaneously addressing sequence length and per-step computation. 
For fairness, both +DEER and +DEER+R-Stitch results are reproduced using the official vLLM-based implementation of DEER, where the full context is re-prefilled after each early-exit decision. This design substantially underestimates the achievable efficiency gains, limiting the apparent acceleration of both DEER alone and our combined method. Our integration retains DEER’s stopping criterion but substitutes its single-model decoding with \methodname, ensuring that the reported improvements are attributable to our method.

\begin{table}[h]
\centering
\caption{
Comparison of decoding strategies on three code generation benchmarks. 
We report accuracy, latency (s/sample), and relative speedup (computed against the corresponding full LLM decoding). 
LLM-7B and LLM-14B denote DeepSeek-R1-Distill-Qwen-7B and DeepSeek-R1-Distill-Qwen-14B, respectively. 
SLM refers to L1-1.5B-Short~\citep{liu2025understanding}, and SpecDec denotes speculative decoding using the corresponding LLM. 
}
\resizebox{0.99\linewidth}{!}{
\begin{tabular}{l|c|ccc|ccc|ccc|ccc}
\toprule
\multirow{2}{*}{\textbf{Method}} & \multirow{2}{*}{$\boldsymbol{\tau}$} 
& \multicolumn{3}{c|}{LiveCodeBench} 
& \multicolumn{3}{c|}{MBPP} 
& \multicolumn{3}{c|}{HumanEval} 
& \multicolumn{3}{c}{\textbf{Average}} \\
\cmidrule{3-14}& & Acc $\uparrow$ & Lat. $\downarrow$ & Spd. $\uparrow$
& Acc $\uparrow$ & Lat. $\downarrow$ & Spd. $\uparrow$
& Acc $\uparrow$ & Lat. $\downarrow$ & Spd. $\uparrow$
& Acc $\uparrow$ & Lat. $\downarrow$ & Spd. $\uparrow$ \\

\midrule
\rowcolor{gray!10}
SLM         & --    & 8.81 & 9.74 & -- & 27.00 & 1.45 & -- & 42.70 & 2.26 & -- & 26.17 & 4.48 & -- \\ 
\rowcolor{gray!15}
LLM-7B      & --    & 40.90 & 92.46 & 1.00$\times$ & 64.00 & 23.97 & 1.00$\times$ & 78.60 & 38.97 & 1.00$\times$ & 61.17 & 51.80 & 1.00$\times$ \\
SpecDec     & --    & 40.31 & 112.21 & 0.82$\times$ & 62.40 & 18.72 & 1.28$\times$ & 80.49 & 47.83 & 0.81$\times$ & 61.07 & 59.59 & 0.87$\times$ \\
R-Stitch    & 0.001 & 40.31 & 87.67 & 1.05$\times$ & 59.30 & 19.62 & 1.22$\times$ & 74.40 & 22.32 & 1.75$\times$ & 58.00 & 43.20 & 1.20$\times$ \\
R-Stitch    & 0.005 & 41.88 & 82.12 & 1.07$\times$ & 48.90 & 13.65 & 1.76$\times$ & 67.70 & 14.18 & 2.75$\times$ & 52.83 & 36.65 & 1.41$\times$ \\
R-Stitch    & 0.01  & 36.59 & 78.54 & 1.18$\times$ & 49.70 & 10.98 & 2.18$\times$ & 62.80 & 7.46  & 5.22$\times$ & 49.70 & 32.33 & 1.60$\times$ \\
\rowcolor{gray!15}
LLM-14B     & --    & 49.51 & 183.26 & 1.00$\times$ & 74.90 & 7.52  & 1.00$\times$ & 86.00 & 25.10 & 1.00$\times$ & 70.14 & 71.96 & 1.00$\times$ \\
SpecDec     & --    & 42.27 & 234.62 & 0.78$\times$ & 72.20 & 4.25 & 1.77$\times$ & 78.60 & 16.92 & 1.48$\times$ & 64.36 & 85.26 & 0.84$\times$ \\
R-Stitch    & 0.001 & 40.90 & 160.10 & 1.14$\times$ & 68.85 & 7.13  & 1.06$\times$ & 77.40 & 16.65 & 1.51$\times$ & 62.38 & 61.29 & 1.17$\times$ \\
R-Stitch    & 0.005 & 36.20 & 142.03 & 1.29$\times$ & 62.40 & 5.83  & 1.29$\times$ & 70.10 & 8.06  & 3.11$\times$ & 56.23 & 51.97 & 1.38$\times$ \\
R-Stitch    & 0.01  & 37.78 & 131.43 & 1.39$\times$ & 56.10 & 4.30  & 1.75$\times$ & 73.20 & 5.86  & 4.28$\times$ & 55.69 & 47.20 & 1.52$\times$ \\
\bottomrule
\end{tabular}
}
\label{tab:code-hybrid-results}
\end{table}

\subsection{Performance on code generation benchmarks.}
Table~\ref{tab:code-hybrid-results} reports accuracy, latency, and relative speedup on programming benchmarks including LiveCodeBench~\citep{jainlivecodemain}, MBPP~\citep{austin2021program}, and HumanEval~\citep{chen2021evaluating}. Compared to math reasoning, acceleration on code generation is more modest: when targeting accuracy comparable to the full LLM, the achievable latency reduction is limited. We attribute this to the characteristics of the specific SLM used in our study, which, while effective on math reasoning tasks, produces less concise and less reliable traces for code. This constrains the gains realizable from hybrid decoding on code benchmarks. Nevertheless, \methodname~still enables a flexible trade-off between efficiency and accuracy across model scales. By adjusting the entropy threshold, one can smoothly interpolate between full LLM accuracy and substantially lower latency, yielding deployment options adapted to different computational budgets and latency requirements.
This shows that, even when the paired SLM is less competitive, token-level hybrid decoding offers a principled and flexible mechanism to balance quality and efficiency, ensuring that practitioners can extract most of the benefits of large models at significantly reduced cost.

\rebuttal{\subsection{Comparison with Same-Family Speculative Decoding}}

\rebuttal{We report results where speculative decoding is evaluated using both the model pairs in the manuscript and the same-family DeepSeek-R1--Distill-Qwen draft--target pairs with higher consistency (marked with *). All results in Table~\ref{tab:same_family_specdec} show accuracy (\%) and latency (s) under an 8k budget. We highlight in \textbf{bold} the best acceleration achieved without any accuracy drop. While same-family pairing improves speculative decoding by increasing draft--target agreement, its acceleration remains limited because strict token-level consistency is still required. R-Stitch achieves comparable or better accuracy with substantially lower latency, as it does not rely on distributional alignment and can exploit concise community SLMs across model families without retraining. Extended comparisons have been added to Appendix A.9 in the revised manuscript.}

{
\color{blue}
\begin{table}[h]
\centering
\small
\caption{\rebuttal{Results of speculative decoding with same-family and manuscript model pairs.}}
\resizebox{0.9\linewidth}{!}{
\begin{tabular}{l|cc|cc|cc|cc}
\toprule

\multirow{2}{*}{\textbf{Method}}
& \multicolumn{2}{c|}{AIME}
& \multicolumn{2}{c|}{AMC}
& \multicolumn{2}{c|}{MATH}
& \multicolumn{2}{c}{Minerva} \\
\cmidrule{2-9}
& Acc $\uparrow$ & Lat. $\downarrow$
& Acc $\uparrow$ & Lat. $\downarrow$
& Acc $\uparrow$ & Lat. $\downarrow$
& Acc $\uparrow$ & Lat. $\downarrow$ \\
\midrule
\rowcolor{gray!15}
LLM 7B           & 33.33 & 86.63   & 66.27 & 63.15   & 86.00 & 38.34   & 31.62 & 41.79 \\
SpecDec 7B       & 36.67 & 201.23  & 69.88 & 95.42   & 87.00 & 48.71   & 34.19 & 56.59 \\
R-Stitch 7B      & \textbf{40.00} & \textbf{62.03}
                 & \textbf{69.88} & \textbf{34.89}
                 & \textbf{87.00} & \textbf{16.61}
                 & \textbf{33.09} & \textbf{18.98} \\
SpecDec 7B*      & 33.33 & 68.65   & 71.08 & 58.53   & 89.20 & 36.86   & 34.19 & 43.33 \\
R-Stitch 7B*     & 36.67 & 69.73   & 68.67 & 61.16   & 89.20 & 40.55   & 36.76 & 41.15 \\
\midrule
\rowcolor{gray!15}
LLM 14B          & 43.33 & 153.20  & 68.67 & 101.76  & 86.00 & 41.66   & 35.29 & 48.02 \\
SpecDec 14B      & 50.00 & 139.10  & 68.67 & 95.59   & 82.80 & 45.88   & 33.46 & 53.02 \\
R-Stitch 14B     & \textbf{43.33} & \textbf{87.61}
                 & \textbf{69.88} & \textbf{41.82}
                 & \textbf{85.20} & \textbf{20.53}
                 & \textbf{35.66} & \textbf{22.16} \\
SpecDec 14B*     & 50.00 & 127.59  & 74.70 & 71.39   & 83.80 & 37.02   & 33.46 & 40.95 \\
R-Stitch 14B*    & 43.33 & 112.23  & 74.70 & 73.24   & 85.20 & 35.41   & 35.66 & 38.89 \\
\bottomrule
\end{tabular}
}
\label{tab:same_family_specdec}
\end{table}
}

\rebuttal{\subsection{Batch Size Analysis}}

\rebuttal{We report here the effect of batch size on R-Stitch latency. Because R-Stitch performs token-level switching between the SLM and LLM, the acceleration ratio naturally decreases as batch size grows; this behavior is shared by speculative-decoding–style methods, where alternating model invocations introduce synchronization points and reduce GPU utilization.}

\rebuttal{We evaluate R-Stitch and speculative decoding on the MATH dataset using the DeepSeek-R1–Distill-Qwen-7B pair within vLLM, with $\tau = 0.02$. The full LLM reaches 86.00\% accuracy, and both acceleration methods obtain 87.00\% accuracy. Table~\ref{tab:batch_size_results} reports average per-sample latency under each batch size. Although acceleration benefits diminish at larger batches, R-Stitch consistently achieves lower latency than speculative decoding, owing to its ability to exploit concise SLM reasoning traces beyond strict token-level consistency.}

\rebuttal{\begin{table}[h]
\centering
\small
\caption{\rebuttal{Latency (s) of the LLM, speculative decoding, and R-Stitch across batch sizes on MATH.}}
\begin{tabular}{lcccccccc}
\toprule
\textbf{Method / Batch} & \textbf{1} & \textbf{2} & \textbf{4} & \textbf{8} & \textbf{16} & \textbf{32} & \textbf{64} & \textbf{128} \\
\midrule
LLM        & 38.34 & 31.58 & 22.07 & 16.11 & 11.08 & 8.42 & 6.52 & 5.97 \\
SpecDec    & 36.86 & 31.61 & 24.29 & 22.04 & 19.20 & 17.88 & 15.32 & 14.01 \\
R-Stitch   & 16.61 & 20.48 & 17.98 & 14.98 & 12.02 & 10.61 & 7.43 & 6.55 \\
\bottomrule
\end{tabular}
\label{tab:batch_size_results}
\end{table}}

\rebuttal{\subsection{Cross-Domain Evaluation}}

\rebuttal{We evaluate R-Stitch ($\tau = 0.02$) and R-Stitch+ beyond mathematical reasoning on four QA and reasoning benchmarks: GPQA-D~\citep{rein2024gpqa}, ZebraLogicBench~\citep{linzebralogic}, CRUXEval~\citep{gucruxeval}, and MMLU-Redux~\citep{gema2025we}. Accuracy (↑) and latency (↓) are reported. For R-Stitch+, the router is trained only on math datasets and directly applied to these domains without any retraining.}

\begin{table}[h]
\centering
\small
\caption{\rebuttal{Cross-domain evaluation on reasoning and QA benchmarks.}}
\resizebox{0.8\linewidth}{!}{
\begin{tabular}{l|cc|cc|cc|cc}
\toprule
\multirow{2}{*}{\textbf{Method}}
& \multicolumn{2}{c|}{GPQA-D}
& \multicolumn{2}{c|}{ZebraLogicBench}
& \multicolumn{2}{c|}{CRUXEval}
& \multicolumn{2}{c}{MMLU-Redux} \\
\cmidrule{2-9}
& Acc $\uparrow$ & Lat. $\downarrow$
& Acc $\uparrow$ & Lat. $\downarrow$
& Acc $\uparrow$ & Lat. $\downarrow$
& Acc $\uparrow$ & Lat. $\downarrow$ \\
\midrule
\rowcolor{gray!15}
LLM-7B     & 23.74 & 171.76 & 21.09 & 91.85 & 70.25 & 14.54 & 74.64 & 20.10 \\
SpecDec    & 24.24 & 152.23 & 20.22 & 72.03 & 70.50 & 14.78 & 74.04 & 19.37 \\
R-Stitch   & 35.86 & 44.57  & 20.78 & 48.23 & 87.29 & 18.23 & 74.04 & 8.12  \\
R-Stitch+  & 34.34 & 39.02  & 20.82 & 45.12 & 69.88 & 9.15  & 74.11 & 8.52  \\

\bottomrule
\end{tabular}
}
\label{tab:cross_domain}
\end{table}

\rebuttal{R-Stitch maintains accuracy comparable to the LLM across all tasks while providing substantial latency reductions. Moreover, R-Stitch+ consistently delivers slightly improved accuracy despite being trained only on math data, showing that its learned policy transfers well across tasks without retraining.}

\rebuttal{\subsection{Architecture and Overhead of Router}}

\rebuttal{The router is explicitly designed to be lightweight, and its computational and memory overhead is negligible. It is a small feed-forward network operating on SLM hidden states, composed of 6 FFN blocks with GELU and residual connections, followed by normalization and a two-head linear layer. The input matches the SLM hidden size (3584 for the 1.5B model), and the total parameter count is 0.17M. On an NVIDIA A100, the router adds 1.05 ms per invocation. At $\tau = 0.02$ on AIME with the 7B pair, it is invoked 350.38 times on average, giving 368 ms total overhead. The full inference time for the same setup is 62.03 s, so the router contributes under 0.6\% of end-to-end latency, demonstrating that its deployability costs are minimal.}

\rebuttal{\subsection{Computational Overhead of Entropy-Based Routing}}

\rebuttal{The cost of entropy evaluation is negligible: entropy is obtained directly from the SLM logits and adds only 0.028 ms per token in our implementation. In the 7B–1.5B AIME setup ($\tau = 0.02$), decoding 4530.60 tokens takes 62.03 seconds, while entropy computation accounts for just 127 ms (approximately 0.2\%), indicating no meaningful runtime burden.}


\rebuttal{\subsection{Comparison with Reward-Guided Speculative Decoding}}

\rebuttal{We compare R-Stitch with Reward-Guided Speculative Decoding (RSD)~\citep{liaoreward}, which performs step-level switching based on an external process reward model (PRM). Since PRMs are trained for specific target models and transfer poorly across different model pairs, RSD’s routing decisions become unstable when the draft–target model combination changes. In contrast, R-Stitch performs fine-grained token-level routing without any external supervision, providing more stable efficiency gains across diverse model pairs. Table~\ref{tab:rsd-comparison} reports accuracy (\%) and latency (s) under an 8k decoding budget.}

\begin{table}[h]
\centering
\small
\caption{\rebuttal{Comparison between RSD and R-Stitch under an 8k decoding budget.}}
\label{tab:rsd-comparison}
\resizebox{0.8\linewidth}{!}{
\begin{tabular}{l|cc|cc|cc|cc}
\toprule
\multirow{2}{*}{\textbf{Method}}
& \multicolumn{2}{c|}{AIME}
& \multicolumn{2}{c|}{AMC}
& \multicolumn{2}{c|}{MATH}
& \multicolumn{2}{c}{Minerva} \\
\cmidrule{2-9}
& Acc $\uparrow$ & Lat. $\downarrow$
& Acc $\uparrow$ & Lat. $\downarrow$
& Acc $\uparrow$ & Lat. $\downarrow$
& Acc $\uparrow$ & Lat. $\downarrow$ \\
\midrule
\rowcolor{gray!15}
LLM-7B        & 33.33 & 86.63  & 66.27 & 63.15  & 86.00 & 38.34 & 31.62 & 41.79 \\
RSD-7B        & 30.00 & 69.73  & 66.27 & 54.12  & 81.60 & 25.12 & 33.80 & 54.37 \\
R-Stitch-7B   & 40.00 & 62.03  & 69.88 & 34.89  & 87.00 & 16.61 & 33.09 & 18.98 \\

\bottomrule
\end{tabular}
}
\end{table}

\subsection{More Per-Sample Comparisons with Speculative Decoding}
\label{app:spec-decoding}

\begin{figure}[h]
    \centering
    \includegraphics[width=0.998\linewidth]{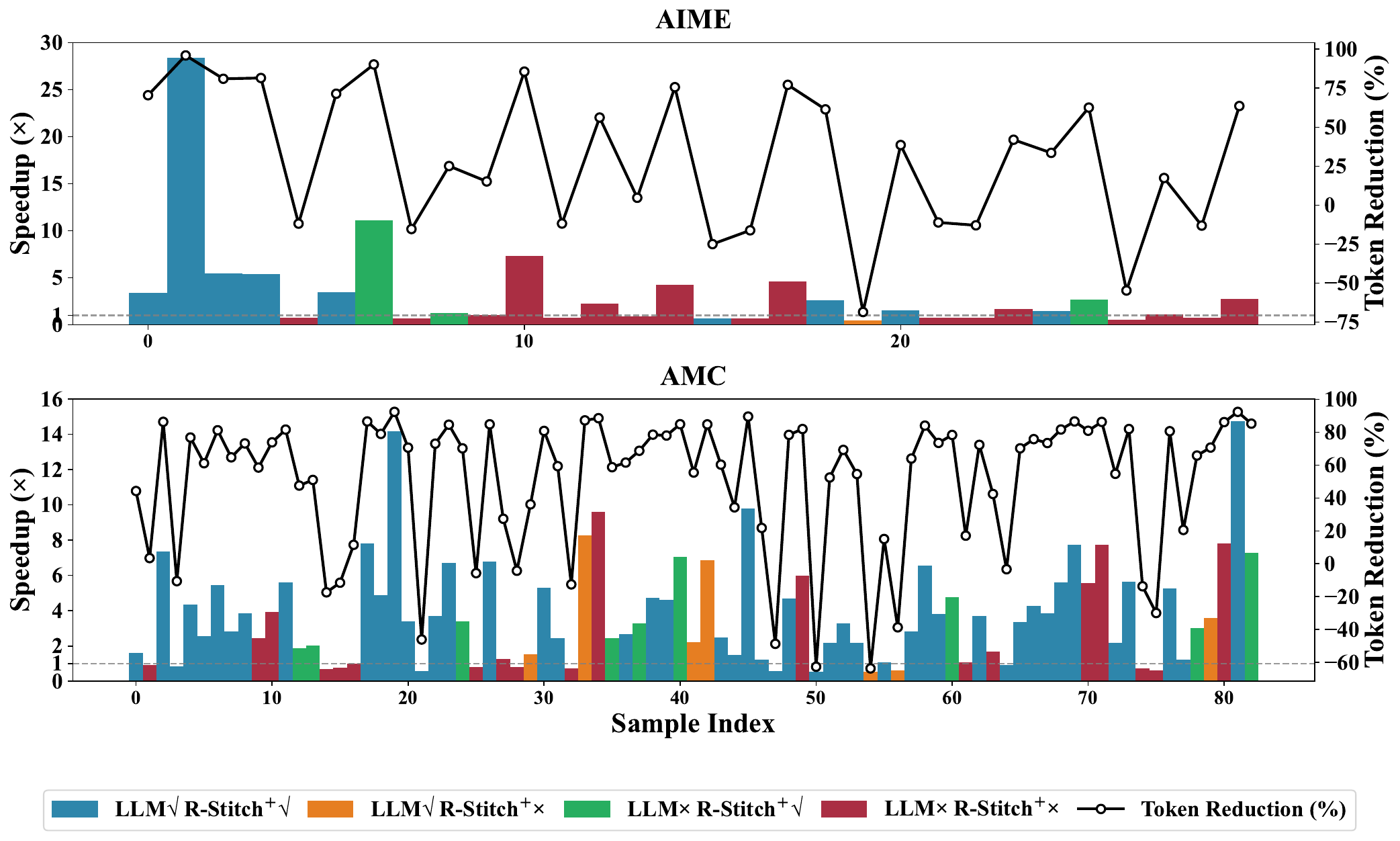}
    \caption{\textbf{Per-sample visualization of R-Stitch$^{+}$ (LLM-7B).}  
    Each bar shows the latency speedup of one sample relative to the baseline LLM.  
    Bar colors encode correctness outcomes of the baseline and R-Stitch$^{+}$.  
    The dashed horizontal line at 1 indicates no speedup.  
    The black solid curve with hollow-circle markers represents token reduction percentages per sample.}
    \label{fig:rstitch-all}
\end{figure}

\begin{figure}[h]
    \centering
    \includegraphics[width=0.998\linewidth]{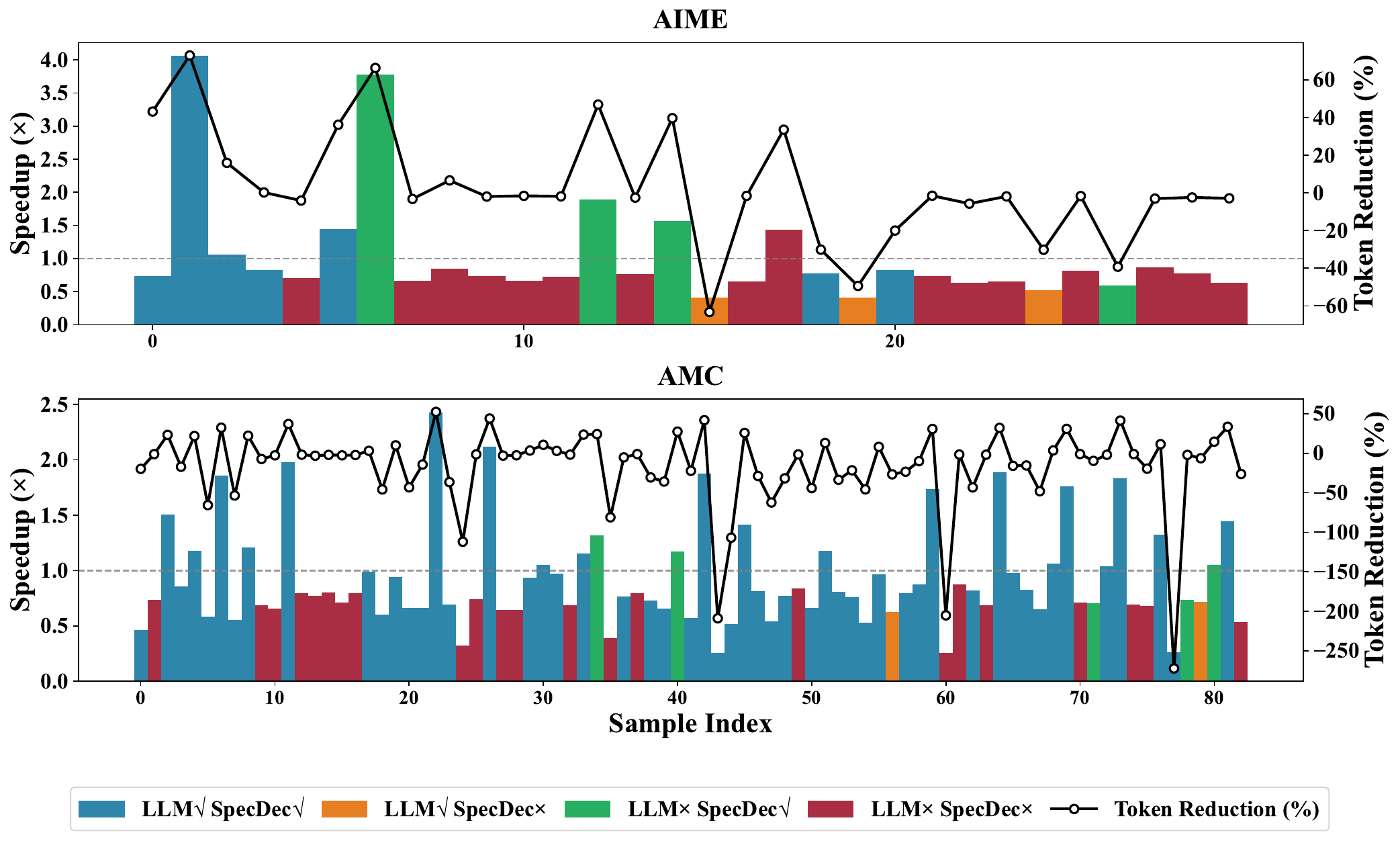}
    \caption{\textbf{Per-sample visualization of Speculative Decoding (LLM-7B).}  
    Each bar shows the latency speedup of one sample relative to the baseline LLM.  
    Bar colors encode correctness outcomes of the baseline and Speculative Decoding.  
    The dashed horizontal line at 1 indicates no speedup.  
    The black solid curve with hollow-circle markers represents token reduction percentages per sample.}
    \label{fig:specdec-all}
\end{figure}

Figures~\ref{fig:rstitch-all} and \ref{fig:specdec-all} present additional per-sample visualizations of latency speedup and token reduction when applying R-Stitch$^{+}$ and Speculative Decoding on the LLM-7B model across more datasets beyond those reported in the main text. The supplementary results confirm the same pattern: R-Stitch$^{+}$ consistently accelerates the majority of samples, whereas Speculative Decoding provides gains only on a minority and often leads to slowdowns due to consistency issues.

\subsection{Additional Results on Token Usage}
In this section, we provide detailed tables for each dataset and benchmark under different decoding budgets as shown in Table~\ref{tab:aime-8k} to Table~\ref{tab:olympiad_bench-16k}. 
Beyond the main results on accuracy, latency, and relative speedup, these tables additionally report the number of tokens consumed by the LLM and the SLM, as well as the total token usage. 
We also compute the percentage of token reduction compared to full LLM decoding. 
These results offer a clearer view of how R-Stitch reduces the reliance on expensive LLM tokens across mathematical reasoning and code generation tasks, while maintaining competitive accuracy and achieving substantial efficiency improvements.

\begin{table}[h]
\centering
\caption{Comparison on \textbf{AIME} with decoding budget = 8k tokens.}
\resizebox{\linewidth}{!}{
\begin{tabular}{l|c|ccc|cccc}
\toprule
\textbf{Method} & $\boldsymbol{\tau}$ & Acc & Lat. & Spd. & LLM Tok. & SLM Tok. & Total Tok. & Reduction (\%)\\
\midrule
SLM & -- & 10.00 & 5.91 & -- & -- & -- & 478.27 & -- \\
\rowcolor{gray!15}
LLM-7B & -- & 33.33 & 86.63 & 1.00$\times$ & -- & -- & 6699.63 & -- \\
SpecDec & -- & 36.67 & 201.23 & 0.43$\times$ & -- & -- & 6922.83 & \textcolor{red}{+3.33\%} \\
R-Stitch & 0.001 & 36.67 & 89.86 & 0.96$\times$ & 3970.03 & 2916.57 & 6886.60 & \textcolor{red}{+2.79\%} \\
R-Stitch & 0.02 & 40.00 & 62.03 & 1.40$\times$ & 2607.24 & 1923.36 & 4530.60 & \textcolor{blue}{-32.38\%} \\
R-Stitch & 0.03 & 30.00 & 42.06 & 2.06$\times$ & 1052.17 & 1914.37 & 2966.54 & \textcolor{blue}{-55.72\%} \\
\rowcolor{gray!15}
LLM-14B & -- & 43.33 & 153.20 & 1.00$\times$ & -- & -- & 6243.13 & -- \\
SpecDec & -- & 50.00 & 139.10 & 1.10$\times$ & -- & -- & 6192.73 & \textcolor{blue}{-0.81\%} \\
R-Stitch & 0.001 & 50.00 & 129.88 & 1.18$\times$ & 3080.07 & 2498.33 & 5578.40 & \textcolor{blue}{-10.65\%} \\
R-Stitch & 0.02 & 43.33 & 87.61 & 1.75$\times$ & 1693.90 & 2386.07 & 4079.97 & \textcolor{blue}{-34.65\%} \\
R-Stitch & 0.03 & 43.33 & 61.59 & 2.49$\times$ & 1030.93 & 2040.03 & 3070.96 & \textcolor{blue}{-50.81\%} \\
\rowcolor{gray!15}
LLM-32B & -- & 43.33 & 354.85 & 1.00$\times$ & -- & -- & 7334.00 & -- \\
SpecDec & -- & 40.00 & 270.25 & 1.31$\times$ & -- & -- & 7540.83 & \textcolor{red}{+2.82\%} \\
R-Stitch & 0.001 & 50.00 & 261.86 & 1.36$\times$ & 4180.43 & 2613.80 & 6794.23 & \textcolor{blue}{-7.36\%} \\
R-Stitch & 0.02 & 50.00 & 184.97 & 1.92$\times$ & 2486.10 & 3085.17 & 5571.27 & \textcolor{blue}{-24.04\%} \\
R-Stitch & 0.03 & 40.00 & 178.19 & 1.99$\times$ & 2284.67 & 3352.27 & 5636.94 & \textcolor{blue}{-23.14\%} \\
\bottomrule
\end{tabular}}
\label{tab:aime-8k}
\end{table}

\begin{table}[h]
\centering
\caption{Comparison on \textbf{AIME} with decoding budget = 16k tokens.}
\resizebox{\linewidth}{!}{
\begin{tabular}{l|c|ccc|cccc}
\toprule
\textbf{Method} & $\boldsymbol{\tau}$ & Acc & Lat. & Spd. & LLM Tok. & SLM Tok. & Total Tok. & Reduction (\%)\\
\midrule
SLM & -- & 10.00 & 5.91 & -- & -- & -- & 478.27 & -- \\
\rowcolor{gray!15}
LLM-7B & -- & 40.00 & 195.77 & 1.00$\times$ & -- & -- & 11739.10 & -- \\
SpecDec & -- & 50.00 & 258.54 & 0.76$\times$ & -- & -- & 12478.27 & \textcolor{red}{+6.30\%} \\
R-Stitch & 0.001 & 46.67 & 192.22 & 1.02$\times$ & 5283.83 & 3898.57 & 9182.40 & \textcolor{blue}{-21.78\%} \\
R-Stitch & 0.02 & 50.00 & 110.16 & 1.78$\times$ & 2325.67 & 3402.33 & 5728.00 & \textcolor{blue}{-51.21\%} \\
R-Stitch & 0.03 & 36.67 & 41.51 & 4.72$\times$ & 920.10 & 1753.27 & 2673.37 & \textcolor{blue}{-77.23\%} \\
R-Stitch$^+$ & -- & 50.00 & 86.03 & 2.28$\times$ & 1911.24 & 2345.13 & 4256.37 & \textcolor{blue}{-63.74\%} \\
\rowcolor{gray!15}
LLM-14B & -- & 50.00 & 246.95 & 1.00$\times$ & -- & -- & 9130.13 & -- \\
SpecDec & -- & 50.00 & 275.64 & 0.90$\times$ & -- & -- & 6312.04 & \textcolor{blue}{-30.87\%} \\
R-Stitch & 0.001 & 56.67 & 217.06 & 1.14$\times$ & 4476.53 & 3550.63 & 8027.16 & \textcolor{blue}{-12.08\%} \\
R-Stitch & 0.02 & 43.33 & 99.12 & 2.49$\times$ & 1746.50 & 2654.30 & 4400.80 & \textcolor{blue}{-51.80\%} \\
R-Stitch & 0.03 & 40.00 & 54.89 & 4.50$\times$ & 983.83 & 1813.63 & 2797.46 & \textcolor{blue}{-69.36\%} \\
R-Stitch$^+$ & -- & 50.00 & 132.98 & 1.86$\times$ & 2784.67 & 2367.77 & 5152.44 & \textcolor{blue}{-43.57\%} \\
\rowcolor{gray!15}
LLM-32B & -- & 56.67 & 591.67 & 1.00$\times$ & -- & -- & 11602.67 & -- \\
SpecDec & -- & 70.00 & 526.03 & 1.12$\times$ & -- & -- & 10879.17 & \textcolor{blue}{-6.24\%} \\
R-Stitch & 0.001 & 70.00 & 488.87 & 1.21$\times$ & 6963.43 & 3908.77 & 10872.20 & \textcolor{blue}{-6.30\%} \\
R-Stitch & 0.02 & 50.00 & 333.17 & 1.78$\times$ & 4119.00 & 4431.53 & 8550.53 & \textcolor{blue}{-26.31\%} \\
R-Stitch & 0.03 & 53.33 & 276.03 & 2.14$\times$ & 3302.57 & 4098.20 & 7400.77 & \textcolor{blue}{-36.21\%} \\
\bottomrule
\end{tabular}}
\label{tab:aime-16k}
\end{table}

\begin{table}[h]
\centering
\caption{Comparison on \textbf{AMC} with decoding budget = 8k tokens.}
\resizebox{\linewidth}{!}{
\begin{tabular}{l|c|ccc|cccc}
\toprule
\textbf{Method} & $\boldsymbol{\tau}$ & Acc & Lat. & Spd. & LLM Tok. & SLM Tok. & Total Tok. & Reduction (\%)\\
\midrule
SLM & -- & 50.60 & 5.37 & -- & -- & -- & 437.29 & -- \\
\rowcolor{gray!15}
LLM-7B & -- & 66.27 & 63.15 & 1.00$\times$ & -- & -- & 4796.87 & -- \\
SpecDec & -- & 69.88 & 95.42 & 0.66$\times$ & -- & -- & 4922.23 & \textcolor{red}{+2.61\%} \\
R-Stitch & 0.001 & 77.11 & 58.72 & 1.08$\times$ & 1868.92 & 1792.31 & 3661.23 & \textcolor{blue}{-23.67\%} \\
R-Stitch & 0.02 & 69.88 & 34.89 & 1.81$\times$ & 1365.77 & 1208.35 & 2574.12 & \textcolor{blue}{-46.34\%} \\
R-Stitch & 0.03 & 69.88 & 24.55 & 2.57$\times$ & 484.27 & 1165.06 & 1649.33 & \textcolor{blue}{-65.62\%} \\
R-Stitch$^+$ & -- & 68.67 & 21.08 & 3.00$\times$ & 468.72 & 987.24 & 1455.96 & \textcolor{blue}{-69.65\%} \\
\rowcolor{gray!15}
LLM-14B & -- & 68.67 & 101.76 & 1.00$\times$ & -- & -- & 4199.61 & -- \\
SpecDec & -- & 68.67 & 95.59 & 1.06$\times$ & -- & -- & 4495.41 & \textcolor{red}{+7.04\%} \\
R-Stitch & 0.001 & 69.88 & 79.54 & 1.28$\times$ & 1799.34 & 1803.77 & 3603.11 & \textcolor{blue}{-14.20\%} \\
R-Stitch & 0.02 & 69.88 & 41.82 & 2.43$\times$ & 756.01 & 1366.87 & 2122.88 & \textcolor{blue}{-49.45\%} \\
R-Stitch & 0.03 & 68.67 & 26.43 & 3.85$\times$ & 453.52 & 984.59 & 1438.11 & \textcolor{blue}{-65.76\%} \\
R-Stitch$^+$ & -- & 73.49 & 49.91 & 2.04$\times$ & 1137.86 & 1135.87 & 2273.73 & \textcolor{blue}{-45.86\%} \\
\rowcolor{gray!15}
LLM-32B & -- & 60.24 & 292.78 & 1.00$\times$ & -- & -- & 6081.42 & -- \\
SpecDec & -- & 59.04 & 209.72 & 1.40$\times$ & -- & -- & 6102.70 & \textcolor{red}{+0.35\%} \\
R-Stitch & 0.001 & 68.67 & 209.88 & 1.39$\times$ & 3094.76 & 2131.43 & 5226.19 & \textcolor{blue}{-14.06\%} \\
R-Stitch & 0.02 & 68.67 & 118.26 & 2.48$\times$ & 1567.72 & 2175.42 & 3743.14 & \textcolor{blue}{-38.45\%} \\
R-Stitch & 0.03 & 69.88 & 86.88 & 3.37$\times$ & 1100.34 & 1911.72 & 3012.06 & \textcolor{blue}{-50.47\%} \\
\bottomrule
\end{tabular}}
\label{tab:amc-8k}
\end{table}

\begin{table}[h]
\centering
\caption{Comparison on \textbf{AMC} with decoding budget = 16k tokens.}
\resizebox{\linewidth}{!}{
\begin{tabular}{l|c|ccc|cccc}
\toprule
\textbf{Method} & $\boldsymbol{\tau}$ & Acc & Lat. & Spd. & LLM Tok. & SLM Tok. & Total Tok. & Reduction (\%)\\
\midrule
SLM & -- & 50.60 & 5.37 & -- & -- & -- & 437.29 & -- \\
\rowcolor{gray!15}
LLM-7B & -- & 71.08 & 116.41 & 1.00$\times$ & -- & -- & 7225.24 & -- \\
SpecDec & -- & 80.72 & 132.45 & 0.88$\times$ & -- & -- & 9211.29 & \textcolor{red}{+27.49\%} \\
R-Stitch & 0.001 & 80.72 & 97.26 & 1.20$\times$ & 2902.64 & 2367.42 & 5270.06 & \textcolor{blue}{-27.06\%} \\
R-Stitch & 0.02 & 67.47 & 54.17 & 2.15$\times$ & 1212.94 & 1830.86 & 3043.80 & \textcolor{blue}{-57.87\%} \\
R-Stitch & 0.03 & 66.27 & 38.07 & 3.06$\times$ & 725.36 & 1505.22 & 2230.58 & \textcolor{blue}{-69.13\%} \\
R-Stitch$^+$ & -- & 77.11 & 56.03 & 2.08$\times$ & 1162.74 & 1632.52 & 2795.26 & \textcolor{blue}{-61.31\%} \\
\rowcolor{gray!15}
LLM-14B & -- & 86.75 & 146.51 & 1.00$\times$ & -- & -- & 5619.14 & -- \\
SpecDec & -- & 83.13 & 138.53 & 1.06$\times$ & -- & -- & 4310.51 & \textcolor{blue}{-23.29\%} \\
R-Stitch & 0.001 & 79.52 & 90.72 & 1.61$\times$ & 1968.23 & 1906.05 & 3874.28 & \textcolor{blue}{-31.05\%} \\
R-Stitch & 0.02 & 66.27 & 51.29 & 2.86$\times$ & 900.52 & 1524.18 & 2424.70 & \textcolor{blue}{-56.85\%} \\
R-Stitch & 0.03 & 63.86 & 28.38 & 5.16$\times$ & 460.84 & 1062.13 & 1522.97 & \textcolor{blue}{-72.90\%} \\
R-Stitch$^+$ & -- & 78.31 & 68.50 & 2.14$\times$ & 1476.96 & 1450.69 & 2927.65 & \textcolor{blue}{-47.90\%} \\
\rowcolor{gray!15}
LLM-32B & -- & 87.95 & 419.94 & 1.00$\times$ & -- & -- & 8384.20 & -- \\
SpecDec & -- & 87.95 & 326.73 & 1.29$\times$ & -- & -- & 8217.63 & \textcolor{blue}{-1.99\%} \\
R-Stitch & 0.001 & 90.36 & 288.92 & 1.45$\times$ & 4251.51 & 2716.18 & 6967.69 & \textcolor{blue}{-16.89\%} \\
R-Stitch & 0.02 & 81.93 & 168.19 & 2.50$\times$ & 2081.64 & 2629.07 & 4710.71 & \textcolor{blue}{-43.81\%} \\
R-Stitch & 0.03 & 73.49 & 143.53 & 2.93$\times$ & 1639.04 & 2661.71 & 4300.75 & \textcolor{blue}{-48.70\%} \\
\bottomrule
\end{tabular}}
\label{tab:amc-16k}
\end{table}

\begin{table}[h]
\centering
\caption{Comparison on \textbf{Minerva} with decoding budget = 8k tokens.}
\resizebox{\linewidth}{!}{
\begin{tabular}{l|c|ccc|cccc}
\toprule
\textbf{Method} & $\boldsymbol{\tau}$ & Acc & Lat. & Spd. & LLM Tok. & SLM Tok. & Total Tok. & Reduction (\%)\\
\midrule
SLM & -- & 25.37 & 5.03 & -- & -- & -- & 408.08 & -- \\
\rowcolor{gray!15}
LLM-7B & -- & 31.62 & 41.79 & 1.00$\times$ & -- & -- & 2966.47 & -- \\
SpecDec & -- & 34.19 & 56.59 & 0.74$\times$ & -- & -- & 3293.91 & \textcolor{red}{+11.04\%} \\
R-Stitch & 0.001 & 34.19 & 38.73 & 1.08$\times$ & 1320.15 & 1141.49 & 2461.64 & \textcolor{blue}{-17.02\%} \\
R-Stitch & 0.02 & 33.09 & 18.98 & 2.20$\times$ & 455.23 & 820.51 & 1275.74 & \textcolor{blue}{-56.99\%} \\
R-Stitch & 0.03 & 33.09 & 17.72 & 2.36$\times$ & 375.65 & 803.61 & 1179.26 & \textcolor{blue}{-60.25\%} \\
R-Stitch$^+$ & -- & 35.29 & 15.33 & 2.73$\times$ & 398.68 & 756.12 & 1154.80 & \textcolor{blue}{-61.07\%} \\
\rowcolor{gray!15}
LLM-14B & -- & 35.29 & 48.02 & 1.00$\times$ & -- & -- & 3348.93 & -- \\
SpecDec & -- & 34.93 & 53.02 & 0.91$\times$ & -- & -- & 3216.78 & \textcolor{blue}{-3.95\%} \\
R-Stitch & 0.001 & 35.66 & 40.54 & 1.18$\times$ & 1014.94 & 925.07 & 1940.01 & \textcolor{blue}{-42.07\%} \\
R-Stitch & 0.02 & 35.66 & 22.16 & 2.17$\times$ & 409.41 & 776.69 & 1186.10 & \textcolor{blue}{-64.58\%} \\
R-Stitch & 0.03 & 34.93 & 17.59 & 2.73$\times$ & 299.85 & 674.68 & 974.53 & \textcolor{blue}{-70.90\%} \\
R-Stitch$^+$ & -- & 35.66 & 26.66 & 1.80$\times$ & 775.71 & 763.89 & 1539.60 & \textcolor{blue}{-54.03\%} \\
\rowcolor{gray!15}
LLM-32B & -- & 41.18 & 231.10 & 1.00$\times$ & -- & -- & 4720.20 & -- \\
SpecDec & -- & 41.91 & 182.28 & 1.27$\times$ & -- & -- & 4724.27 & \textcolor{red}{+0.09\%} \\
R-Stitch & 0.001 & 42.65 & 141.24 & 1.64$\times$ & 2309.25 & 1529.28 & 3838.53 & \textcolor{blue}{-18.68\%} \\
R-Stitch & 0.02 & 36.76 & 59.58 & 3.88$\times$ & 831.41 & 1176.50 & 2007.91 & \textcolor{blue}{-57.46\%} \\
R-Stitch & 0.03 & 34.56 & 43.41 & 5.32$\times$ & 578.15 & 1002.49 & 1580.64 & \textcolor{blue}{-66.51\%} \\
\bottomrule
\end{tabular}}
\label{tab:minerva-8k}
\end{table}

\begin{table}[h]
\centering
\caption{Comparison on \textbf{Minerva} with decoding budget = 16k tokens.}
\resizebox{\linewidth}{!}{
\begin{tabular}{l|c|ccc|cccc}
\toprule
\textbf{Method} & $\boldsymbol{\tau}$ & Acc & Lat. & Spd. & LLM Tok. & SLM Tok. & Total Tok. & Reduction (\%)\\
\midrule
SLM & -- & 25.37 & 5.03 & -- & -- & -- & 408.08 & -- \\
\rowcolor{gray!15}
LLM-7B & -- & 34.93 & 54.43 & 1.00$\times$ & -- & -- & 3616.44 & -- \\
SpecDec & -- & 35.66 & 115.76 & 0.47$\times$ & -- & -- & 5987.29 & \textcolor{red}{+65.56\%} \\
R-Stitch & 0.001 & 35.66 & 45.53 & 1.20$\times$ & 1482.54 & 1272.18 & 2754.72 & \textcolor{blue}{-23.83\%} \\
R-Stitch & 0.02 & 33.09 & 18.79 & 2.90$\times$ & 443.68 & 810.42 & 1254.10 & \textcolor{blue}{-65.32\%} \\
R-Stitch & 0.03 & 35.29 & 13.50 & 4.03$\times$ & 295.17 & 663.07 & 958.24 & \textcolor{blue}{-73.50\%} \\
R-Stitch$^+$ & -- & 35.29 & 13.68 & 3.98$\times$ & 283.08 & 667.68 & 950.76 & \textcolor{blue}{-73.71\%} \\
\rowcolor{gray!15}
LLM-14B & -- & 39.34 & 68.99 & 1.00$\times$ & -- & -- & 2801.30 & -- \\
SpecDec & -- & 39.34 & 55.40 & 1.25$\times$ & -- & -- & 3521.18 & \textcolor{red}{+25.70\%} \\
R-Stitch & 0.001 & 37.87 & 40.46 & 1.71$\times$ & 990.31 & 924.97 & 1915.28 & \textcolor{blue}{-31.63\%} \\
R-Stitch & 0.02 & 31.99 & 22.74 & 3.03$\times$ & 416.56 & 784.71 & 1201.27 & \textcolor{blue}{-57.12\%} \\
R-Stitch & 0.03 & 33.82 & 16.14 & 4.27$\times$ & 263.77 & 645.72 & 909.49 & \textcolor{blue}{-67.53\%} \\
R-Stitch$^+$ & -- & 37.13 & 35.11 & 1.96$\times$ & 810.15 & 810.38 & 1620.53 & \textcolor{blue}{-42.15\%} \\
\rowcolor{gray!15}
LLM-32B & -- & 46.69 & 348.48 & 1.00$\times$ & -- & -- & 5553.52 & -- \\
SpecDec & -- & 44.23 & 273.07 & 1.28$\times$ & -- & -- & 5982.18 & \textcolor{red}{+7.72\%} \\
R-Stitch & 0.001 & 41.18 & 126.62 & 2.75$\times$ & 2013.80 & 1605.40 & 3619.20 & \textcolor{blue}{-34.83\%} \\
R-Stitch & 0.02 & 40.07 & 73.32 & 4.75$\times$ & 900.55 & 1923.00 & 2823.55 & \textcolor{blue}{-49.16\%} \\
R-Stitch & 0.03 & 36.40 & 43.58 & 8.00$\times$ & 302.19 & 626.28 & 928.47 & \textcolor{blue}{-83.28\%} \\
\bottomrule
\end{tabular}}
\label{tab:minerva-16k}
\end{table}

\begin{table}[h]
\centering
\caption{Comparison on \textbf{MATH} with decoding budget = 8k tokens.}
\resizebox{\linewidth}{!}{
\begin{tabular}{l|c|ccc|cccc}
\toprule
\textbf{Method} & $\boldsymbol{\tau}$ & Acc & Lat. & Spd. & LLM Tok. & SLM Tok. & Total Tok. & Reduction (\%)\\
\midrule
SLM & -- & 73.60 & 4.56 & -- & -- & -- & 370.67 & -- \\
\rowcolor{gray!15}
LLM-7B & -- & 86.00 & 38.34 & 1.00$\times$ & -- & -- & 2753.01 & -- \\
SpecDec & -- & 87.00 & 48.71 & 0.79$\times$ & -- & -- & 3048.78 & \textcolor{red}{+10.74\%} \\
R-Stitch & 0.001 & 89.40 & 32.94 & 1.16$\times$ & 1022.38 & 1101.67 & 2124.05 & \textcolor{blue}{-22.85\%} \\
R-Stitch & 0.02 & 87.00 & 16.61 & 2.31$\times$ & 344.28 & 770.52 & 1114.80 & \textcolor{blue}{-59.51\%} \\
R-Stitch & 0.03 & 85.60 & 15.15 & 2.53$\times$ & 280.26 & 748.21 & 1028.47 & \textcolor{blue}{-62.64\%} \\
R-Stitch$^+$ & -- & 86.60 & 15.68 & 2.45$\times$ & 503.17 & 598.88 & 1102.05 & \textcolor{blue}{-59.97\%} \\
\rowcolor{gray!15}
LLM-14B & -- & 86.00 & 41.66 & 1.00$\times$ & -- & -- & 2934.52 & -- \\
SpecDec & -- & 82.80 & 45.88 & 0.91$\times$ & -- & -- & 2443.78 & \textcolor{blue}{-16.72\%} \\
R-Stitch & 0.001 & 89.00 & 37.03 & 1.13$\times$ & 867.72 & 937.65 & 1805.37 & \textcolor{blue}{-38.48\%} \\
R-Stitch & 0.02 & 85.20 & 20.53 & 2.03$\times$ & 349.82 & 749.75 & 1099.57 & \textcolor{blue}{-62.53\%} \\
R-Stitch & 0.03 & 83.40 & 14.77 & 2.82$\times$ & 215.10 & 645.26 & 860.36 & \textcolor{blue}{-70.68\%} \\
R-Stitch$^+$ & -- & 88.20 & 29.08 & 1.43$\times$ & 635.27 & 764.58 & 1399.85 & \textcolor{blue}{-52.30\%} \\
\rowcolor{gray!15}
LLM-32B & -- & 87.20 & 178.38 & 1.00$\times$ & -- & -- & 3054.16 & -- \\
SpecDec & -- & 88.60 & 136.73 & 1.30$\times$ & -- & -- & 3052.70 & \textcolor{blue}{-0.05\%} \\
R-Stitch & 0.001 & 91.20 & 95.49 & 1.87$\times$ & 1507.99 & 1322.78 & 2830.77 & \textcolor{blue}{-7.31\%} \\
R-Stitch & 0.02 & 90.60 & 80.21 & 2.22$\times$ & 1087.60 & 1209.95 & 2297.55 & \textcolor{blue}{-24.77\%} \\
R-Stitch & 0.03 & 87.00 & 32.21 & 5.54$\times$ & 387.09 & 890.72 & 1277.81 & \textcolor{blue}{-58.16\%} \\
\bottomrule
\end{tabular}}
\label{tab:math-8k}
\end{table}

\begin{table}[h]
\centering
\caption{Comparison on \textbf{MATH} with decoding budget = 16k tokens.}
\resizebox{\linewidth}{!}{
\begin{tabular}{l|c|ccc|cccc}
\toprule
\textbf{Method} & $\boldsymbol{\tau}$ & Acc & Lat. & Spd. & LLM Tok. & SLM Tok. & Total Tok. & Reduction (\%)\\
\midrule
SLM & -- & 73.60 & 4.56 & -- & -- & -- & 370.67 & -- \\
\rowcolor{gray!15}
LLM-7B & -- & 90.80 & 50.07 & 1.00$\times$ & -- & -- & 3381.51 & -- \\
SpecDec & -- & 91.20 & 44.47 & 1.13$\times$ & -- & -- & 2645.27 & \textcolor{blue}{-21.77\%} \\
R-Stitch & 0.001 & 91.00 & 38.34 & 1.31$\times$ & 1136.11 & 1151.15 & 2287.26 & \textcolor{blue}{-32.36\%} \\
R-Stitch & 0.02 & 88.60 & 22.56 & 2.22$\times$ & 451.49 & 908.54 & 1360.03 & \textcolor{blue}{-59.78\%} \\
R-Stitch & 0.03 & 83.20 & 15.62 & 3.21$\times$ & 275.91 & 729.55 & 1005.46 & \textcolor{blue}{-70.27\%} \\
R-Stitch$^+$ & -- & 88.60 & 16.02 & 3.13$\times$ & 278.17 & 732.16 & 1010.33 & \textcolor{blue}{-70.12\%} \\
\rowcolor{gray!15}
LLM-14B & -- & 88.60 & 66.54 & 1.00$\times$ & -- & -- & 2698.10 & -- \\
SpecDec & -- & 87.40 & 67.18 & 0.99$\times$ & -- & -- & 2681.58 & \textcolor{blue}{-0.61\%} \\
R-Stitch & 0.001 & 89.40 & 39.82 & 1.67$\times$ & 886.11 & 958.29 & 1844.40 & \textcolor{blue}{-31.64\%} \\
R-Stitch & 0.02 & 87.80 & 22.62 & 2.94$\times$ & 365.23 & 793.84 & 1159.07 & \textcolor{blue}{-57.04\%} \\
R-Stitch & 0.03 & 84.80 & 17.37 & 3.83$\times$ & 237.71 & 687.69 & 925.40 & \textcolor{blue}{-65.70\%} \\
R-Stitch$^+$ & -- & 89.60 & 33.70 & 1.97$\times$ & 719.47 & 811.61 & 1531.08 & \textcolor{blue}{-43.25\%} \\
\rowcolor{gray!15}
LLM-32B & -- & 94.00 & 249.92 & 1.00$\times$ & -- & -- & 4304.52 & -- \\
SpecDec & -- & 93.20 & 187.79 & 1.33$\times$ & -- & -- & 4023.25 & \textcolor{blue}{-6.53\%} \\
R-Stitch & 0.001 & 94.00 & 172.08 & 1.45$\times$ & 952.53 & 905.83 & 1858.36 & \textcolor{blue}{-56.83\%} \\
R-Stitch & 0.02 & 90.60 & 80.21 & 3.12$\times$ & 1087.60 & 1209.95 & 2297.55 & \textcolor{blue}{-46.62\%} \\
R-Stitch & 0.03 & 87.80 & 38.32 & 6.52$\times$ & 148.40 & 467.67 & 616.07 & \textcolor{blue}{-85.69\%} \\
\bottomrule
\end{tabular}}
\label{tab:math-16k}
\end{table}

\begin{table}[h]
\centering
\caption{Comparison on \textbf{OlympiadBench} with decoding budget = 8k tokens.}
\resizebox{\linewidth}{!}{
\begin{tabular}{l|c|ccc|cccc}
\toprule
\textbf{Method} & $\boldsymbol{\tau}$ & Acc & Lat. & Spd. & LLM Tok. & SLM Tok. & Total Tok. & Reduction (\%)\\
\midrule
SLM & -- & 36.89 & 5.42 & -- & -- & -- & 441.54 & -- \\
\rowcolor{gray!15}
LLM-7B & -- & 51.85 & 117.31 & 1.00$\times$ & -- & -- & 5109.47 & -- \\
SpecDec & -- & 51.85 & 134.23 & 0.87$\times$ & -- & -- & 5292.34 & \textcolor{red}{+3.58\%} \\
R-Stitch & 0.001 & 55.26 & 105.29 & 1.11$\times$ & 1551.23 & 1216.72 & 2767.95 & \textcolor{blue}{-45.83\%} \\
R-Stitch & 0.02 & 51.85 & 70.43 & 1.67$\times$ & 487.93 & 810.37 & 1298.30 & \textcolor{blue}{-74.59\%} \\
R-Stitch & 0.03 & 48.59 & 51.43 & 2.28$\times$ & 280.62 & 647.15 & 927.77 & \textcolor{blue}{-81.84\%} \\
R-Stitch$^+$ & -- & 52.00 & 45.02 & 2.61$\times$ & 374.52 & 679.23 & 1053.75 & \textcolor{blue}{-79.38\%} \\
\rowcolor{gray!15}
LLM-14B & -- & 54.07 & 190.89 & 1.00$\times$ & -- & -- & 4623.48 & -- \\
SpecDec & -- & 54.22 & 180.03 & 1.06$\times$ & -- & -- & 5320.88 & \textcolor{red}{+15.08\%} \\
R-Stitch & 0.001 & 54.22 & 165.07 & 1.16$\times$ & 1285.81 & 938.74 & 2224.55 & \textcolor{blue}{-51.89\%} \\
R-Stitch & 0.02 & 52.44 & 96.09 & 1.99$\times$ & 375.46 & 668.54 & 1044.00 & \textcolor{blue}{-77.42\%} \\
R-Stitch & 0.03 & 48.44 & 52.36 & 3.65$\times$ & 214.74 & 556.72 & 771.46 & \textcolor{blue}{-83.31\%} \\
R-Stitch$^+$ & -- & 56.30 & 108.67 & 1.76$\times$ & 906.33 & 809.67 & 1716.00 & \textcolor{blue}{-62.89\%} \\
\rowcolor{gray!15}
LLM-32B & -- & 50.67 & 379.52 & 1.00$\times$ & -- & -- & 5988.08 & -- \\
SpecDec & -- & 50.37 & 351.27 & 1.08$\times$ & -- & -- & 4564.46 & \textcolor{blue}{-23.77\%} \\
R-Stitch & 0.001 & 50.67 & 341.67 & 1.11$\times$ & 2441.57 & 1432.92 & 3874.49 & \textcolor{blue}{-35.30\%} \\
R-Stitch & 0.02 & 53.78 & 226.71 & 1.67$\times$ & 488.88 & 781.20 & 1270.08 & \textcolor{blue}{-78.79\%} \\
R-Stitch & 0.03 & 52.15 & 157.47 & 2.41$\times$ & 515.87 & 968.86 & 1484.73 & \textcolor{blue}{-75.21\%} \\
\bottomrule
\end{tabular}}
\label{tab:olympiad_bench-8k}
\end{table}

\begin{table}[h]
\centering
\caption{Comparison on \textbf{OlympiadBench} with decoding budget = 16k tokens.}
\resizebox{\linewidth}{!}{
\begin{tabular}{l|c|ccc|cccc}
\toprule
\textbf{Method} & $\boldsymbol{\tau}$ & Acc & Lat. & Spd. & LLM Tok. & SLM Tok. & Total Tok. & Reduction (\%)\\
\midrule
SLM & -- & 36.89 & 5.42 & -- & -- & -- & 441.54 & -- \\
\rowcolor{gray!15}
LLM-7B & -- & 58.67 & 208.77 & 1.00$\times$ & -- & -- & 6839.96 & -- \\
SpecDec & -- & 60.15 & 267.03 & 0.78$\times$ & -- & -- & 7208.65 & \textcolor{red}{+5.39\%} \\
R-Stitch & 0.001 & 60.15 & 201.13 & 1.04$\times$ & 1817.81 & 1327.43 & 3145.24 & \textcolor{blue}{-54.02\%} \\
R-Stitch & 0.02 & 58.67 & 109.98 & 1.90$\times$ & 504.00 & 840.55 & 1344.55 & \textcolor{blue}{-80.34\%} \\
R-Stitch & 0.03 & 54.22 & 65.65 & 3.18$\times$ & 288.84 & 664.50 & 953.34 & \textcolor{blue}{-86.06\%} \\
R-Stitch$^+$ & -- & 59.11 & 95.31 & 2.19$\times$ & 498.17 & 814.36 & 1312.53 & \textcolor{blue}{-80.81\%} \\
\rowcolor{gray!15}
LLM-14B & -- & 60.00 & 316.06 & 1.00$\times$ & -- & -- & 6024.14 & -- \\
SpecDec & -- & 60.00 & 360.14 & 0.88$\times$ & -- & -- & 6143.46 & \textcolor{red}{+1.98\%} \\
R-Stitch & 0.001 & 61.04 & 69.28 & 4.56$\times$ & 1516.14 & 1225.07 & 2741.21 & \textcolor{blue}{-54.50\%} \\
R-Stitch & 0.02 & 54.22 & 108.61 & 2.91$\times$ & 449.36 & 771.67 & 1221.03 & \textcolor{blue}{-79.73\%} \\
R-Stitch & 0.03 & 48.59 & 68.63 & 4.61$\times$ & 257.66 & 623.88 & 881.54 & \textcolor{blue}{-85.37\%} \\
R-Stitch$^+$ & -- & 59.11 & 121.39 & 2.60$\times$ & 982.06 & 846.84 & 1828.90 & \textcolor{blue}{-69.64\%} \\
\rowcolor{gray!15}
LLM-32B & -- & 67.70 & 722.77 & 1.00$\times$ & -- & -- & 8524.71 & -- \\
SpecDec & -- & 67.70 & 798.23 & 0.91$\times$ & -- & -- & 7495.21 & \textcolor{blue}{-12.08\%} \\
R-Stitch & 0.001 & 66.07 & 654.24 & 1.10$\times$ & 6415.09 & 2689.76 & 9104.85 & \textcolor{red}{+6.81\%} \\
R-Stitch & 0.02 & 61.12 & 358.87 & 2.01$\times$ & 3779.95 & 2954.05 & 6734.00 & \textcolor{blue}{-21.01\%} \\
R-Stitch & 0.03 & 55.70 & 209.96 & 3.44$\times$ & 625.54 & 1085.62 & 1711.16 & \textcolor{blue}{-79.93\%} \\
\bottomrule
\end{tabular}}
\label{tab:olympiad_bench-16k}
\end{table}

\subsection{Qualitative Case Studies}
\label{app:qualitative}
To provide a more intuitive understanding of how \methodname\ leverages the complementary strengths of the SLM and LLM, 
we visualize several representative examples. For each selected problem, we compare the responses of the SLM, the LLM, speculative decoding, and our method \methodname\ with $\tau=0.02$. 
For long outputs, we omit intermediate tokens for readability. 

\begin{figure}[h]
    \centering
    \includegraphics[width=0.92\linewidth]{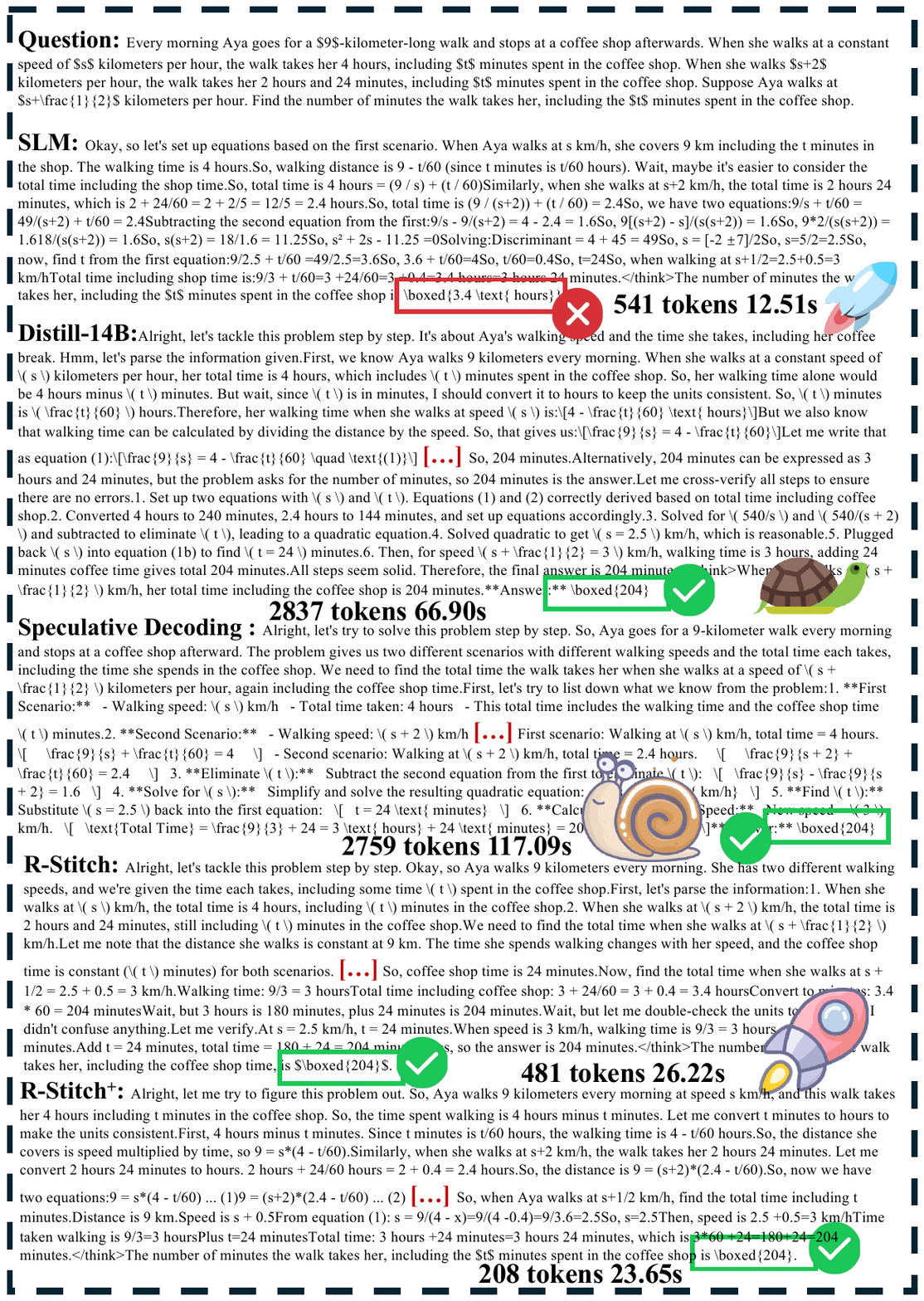}
    \caption{\textbf{Qualitative comparison of outputs.} 
    Each case shows the predicted solution, token count, and latency for the SLM, LLM, speculative decoding, and \methodname\ ($\tau=0.02$). 
    \methodname\ produces concise and correct reasoning while reducing token count and latency compared to LLM decoding.}
    \label{fig:qualitative_vis1}
\end{figure}

\begin{figure}[h]
    \centering
    \includegraphics[width=0.92\linewidth]{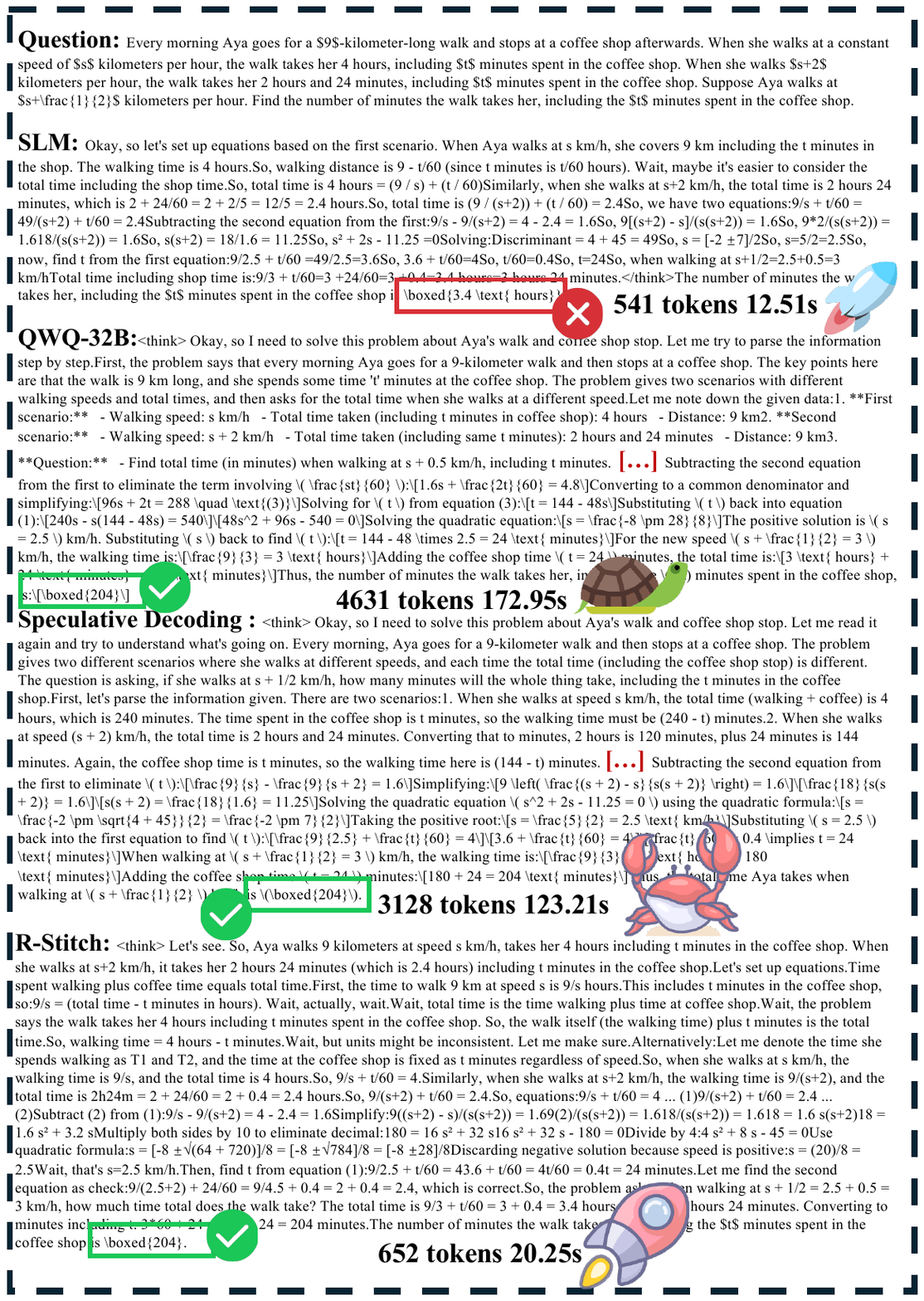}
    \caption{\textbf{Qualitative comparison of outputs.} 
    Each case shows the predicted solution, token count, and latency for the SLM, LLM, speculative decoding, and \methodname\ ($\tau=0.02$). 
    \methodname\ produces concise and correct reasoning while reducing token count and latency compared to LLM decoding.}
    \label{fig:qualitative_vis2}
\end{figure}

\begin{figure}[h]
    \centering
    \includegraphics[width=0.92\linewidth]{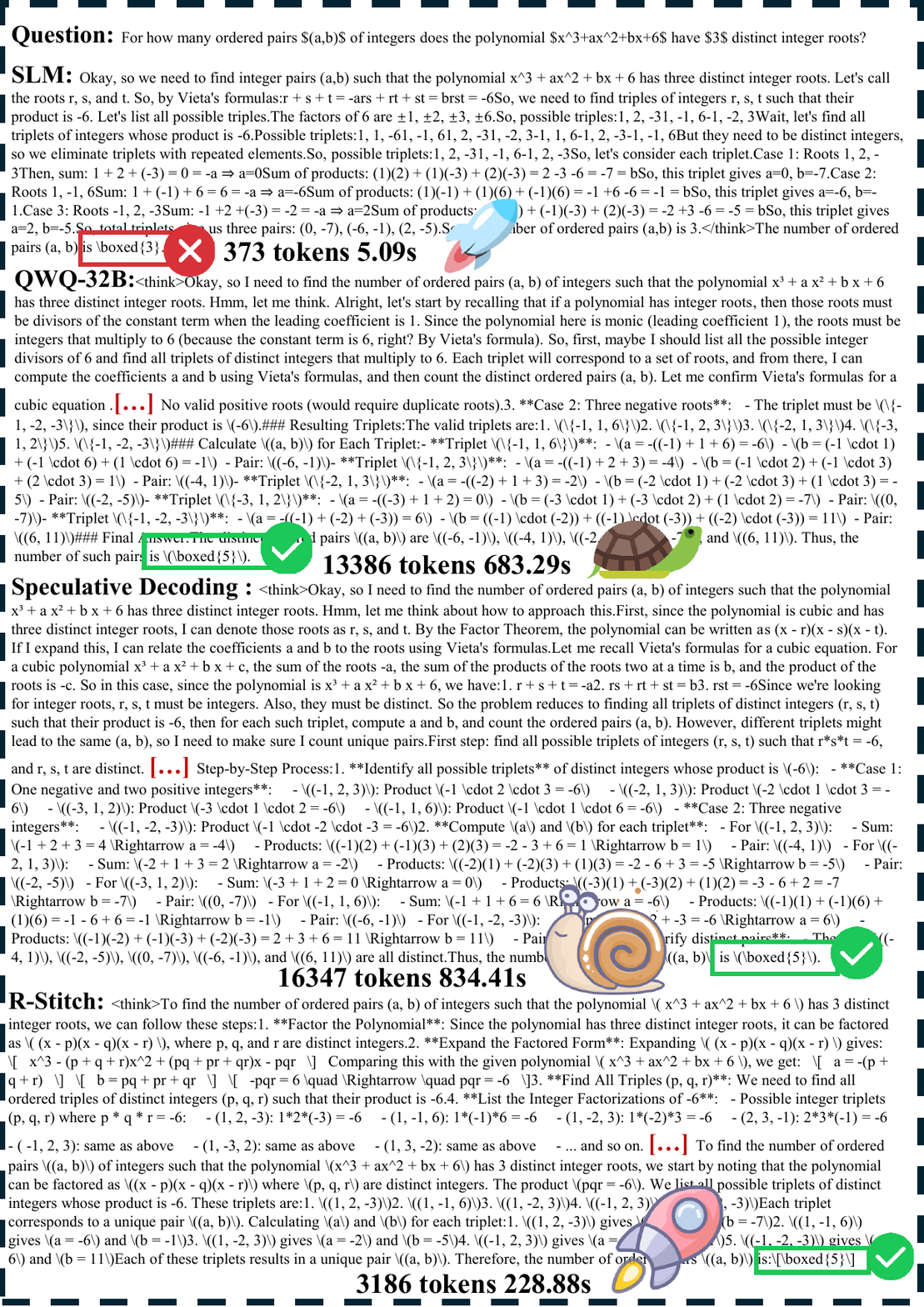}
    \caption{\textbf{Qualitative comparison of outputs.} 
    Each case shows the predicted solution, token count, and latency for the SLM, LLM, speculative decoding, and \methodname\ ($\tau=0.02$). 
    \methodname\ produces concise and correct reasoning while reducing token count and latency compared to LLM decoding.}
    \label{fig:qualitative_vis3}
\end{figure}

\begin{figure}[h]
    \centering
    \includegraphics[width=0.92\linewidth]{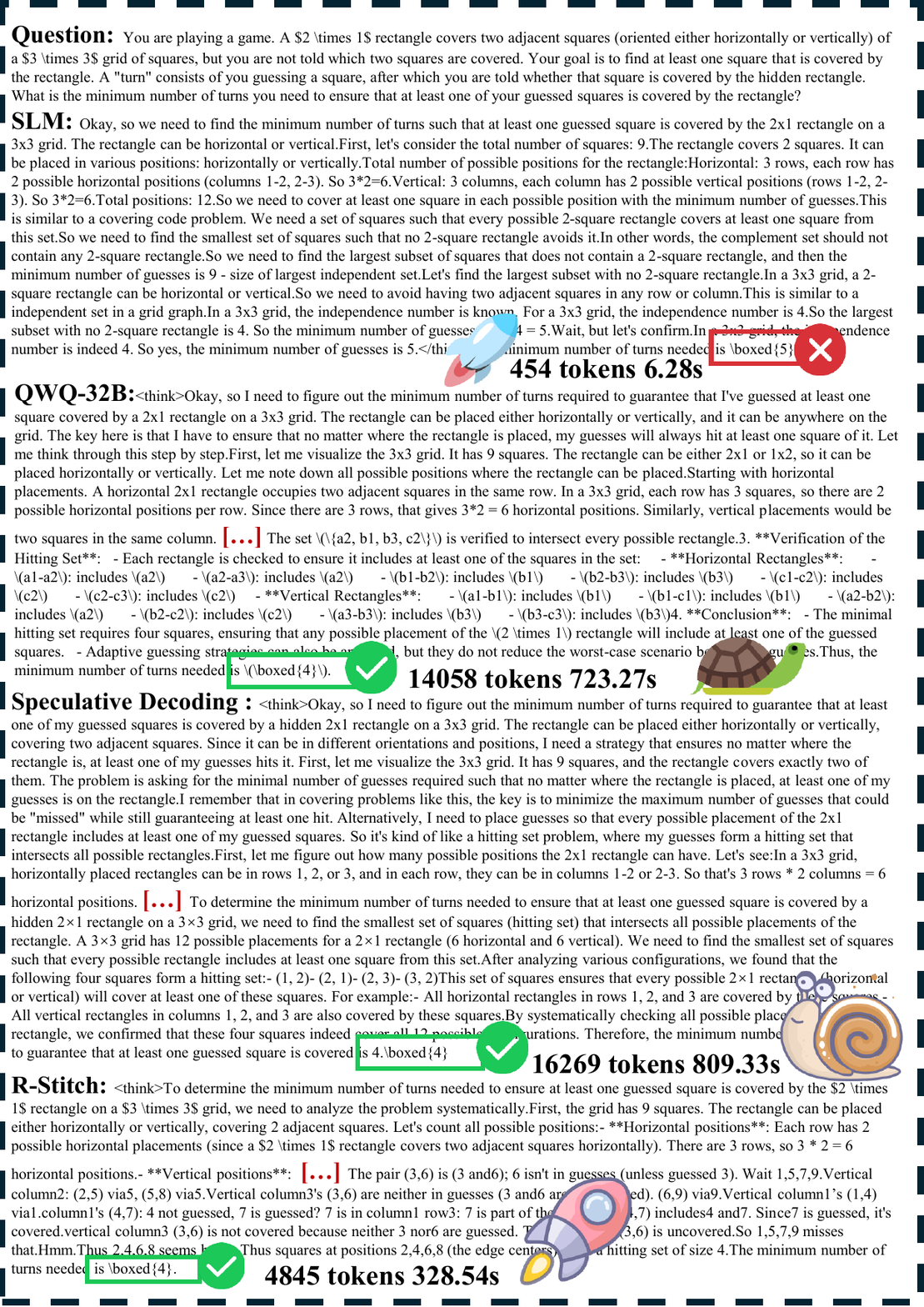}
    \caption{\textbf{Qualitative comparison of outputs.} 
    Each case shows the predicted solution, token count, and latency for the SLM, LLM, speculative decoding, and \methodname\ ($\tau=0.02$). 
    \methodname\ produces concise and correct reasoning while reducing token count and latency compared to LLM decoding.}
    \label{fig:qualitative_vis4}
\end{figure}

As shown in Figure~\ref{fig:qualitative_vis1} to Figure~\ref{fig:qualitative_vis4}, the SLM along generates very concise answers with extremely low latency but often produces incorrect solutions. 
The LLM eventually produces correct answers but requires thousands of tokens and much higher latency. 
Speculative decoding sometimes suffers from frequent token rejections under low model agreement, resulting in more decoding steps than vanilla LLM decoding and hence longer latency. 
In contrast, \methodname\ effectively combines the SLM's brevity with the LLM's accuracy: it selectively accepts confident SLM tokens while routing uncertain ones to the LLM, producing shorter but still correct solutions. 
This substantially reduces the number of generated tokens and yields significant acceleration.


\end{document}